\documentclass{ieeetj}
\usepackage{cite}
\usepackage{amsmath,amssymb,amsfonts}
\usepackage{algorithmic}
\usepackage{color}
\usepackage{graphicx}
\usepackage{textcomp}
\usepackage{xcolor}
\usepackage{hyperref}
\hypersetup{hidelinks=true}
\usepackage{algorithm,algorithmic}
\def\BibTeX{{\rm B\kern-.05em{\sc i\kern-.025em b}\kern-.08em
    T\kern-.1667em\lower.7ex\hbox{E}\kern-.125emX}}
\AtBeginDocument{\definecolor{tmlcncolor}{cmyk}{0.93,0.59,0.15,0.02}\definecolor{NavyBlue}{RGB}{0,86,125}}

\def\OJlogo{\vspace{-4pt}$<$Society logo(s) and publication title will appear here.$>$}
\def\seclogo{\vspace{10pt}$<$Society logo(s) and publication title will appear here.$>$}

\usepackage[whole]{bxcjkjatype}

\usepackage{url}

\usepackage{cleveref}
\crefname{section}{Section}{Section}
\crefname{figure}{Figure}{Figure}
\crefname{subfigure}{Figure}{Figure}
\crefname{table}{Table}{Table}
\crefname{subtable}{Table}{Table}
\crefname{appendix}{Appendix}{Appendix}

\usepackage{setspace}
\usepackage{booktabs}
\usepackage{multirow}
\usepackage[table]{xcolor}
\usepackage[dvipsnames]{xcolor}
\usepackage{pifont}
\newcommand{\gray}[1]{\cellcolor{gray!15}#1}
\newcommand{\heat}[2]{\cellcolor{Cyan!#1}#2}
\newcommand{\improve}[2]{\cellcolor{Cyan!#1}#2\textsuperscript{\textcolor{Cyan!60!black}{\scriptsize +}}}
\newcommand{\degrade}[2]{\cellcolor{red!#1}#2\textsuperscript{\textcolor{red!60!black}{\scriptsize --}}}
\usepackage[caption=false,font=footnotesize]{subfig}

\begin{document}
\receiveddate{XX Month, XXXX}
\reviseddate{XX Month, XXXX}
\accepteddate{XX Month, XXXX}
\publisheddate{XX Month, XXXX}
\currentdate{XX Month, XXXX}
\doiinfo{XXXX.2022.1234567}

\markboth{M3-DuplexBench: A Multi-Turn, Multilingual, Multidomain Benchmark for Full-Duplex Spoken Dialogue Models}{Fukuda {et al.}}
% KOMATSU ET AL.: SPEAKER-DISENTANGLED CHUNK-WISE REGRESSION FOR SYLLABIC TOKENIZATION

\title{M3-DuplexBench: A Multi-Turn, Multilingual, Multidomain Benchmark for Full-Duplex Spoken Dialogue Models}

\author{Ryo Fukuda, Member, IEEE, Atsushi Ando, Hiroki Kanagawa,\\Takatomo Kano, Member, IEEE, Marc Delcroix, Senior Member, IEEE,\\Naohiro Tawara, Member, IEEE, Yuya Chiba}
\affil{NTT, Inc., Japan}
\corresp{Corresponding author: Ryo Fukuda (email: ryo.fukuda@ntt.com).}
\begin{abstract}
Full-duplex spoken dialogue systems (FDSDSs) can listen while speaking, enabling natural conversational behaviors such as smooth turn-taking, backchannel handling, and user barge-in handling.
% However, fair comparisons in multi-turn conversations remain a challenge.
% In addition, existing benchmarks provide limited coverage of languages and dialogue domains.
However, controlled comparisons of FDSDSs in multi-turn interaction remain challenging, and existing benchmarks offer limited joint coverage of multiple languages, dialogue settings, interaction timing, and response content.
We propose M3-DuplexBench, a multi-turn, multilingual, multidomain benchmark for controlled evaluation of FDSDSs.
The benchmark jointly evaluates interaction timing and response content in English and Japanese across casual conversation and multi-turn question answering.
The English and Japanese question-answering data are derived from a common source corpus, enabling cross-lingual comparisons of timing and content performance.
We also provide a reference-conditioned multi-turn setting as an optional diagnostic for examining sensitivity to dialogue history and trajectory mismatch.
Experiments with diverse FDSDSs reveal model-specific interaction patterns, differences across languages and domains, and model-dependent effects of dialogue context on both timing and content.
% We will release the benchmark upon publication of the paper. % OJSP
The benchmark is available at \url{https://github.com/nttcslab-sp/m3-duplexbench}. % arXiv
% The benchmark is available at this https URL (\url{https://github.com/nttcslab-sp/m3-duplexbench}).
\end{abstract}



\maketitle

\section{INTRODUCTION}
\IEEEPARstart{S}{poken} dialogue systems (SDSs) communicate with users through speech.
Conventional SDSs often follow a half-duplex interaction pattern, waiting for the user to finish speaking before responding~\cite{Chen2025-wa,Skantze2021-dg}.
In contrast, full-duplex spoken dialogue systems (FDSDSs), which can listen while speaking, have recently attracted attention~\cite{Nguyen2022-um,hu25f_interspeech}.
This capability supports conversational behaviors such as handling user interruptions, continuing to speak during user backchannels, and producing natural overlapping speech.

Along with the development of FDSDSs, many benchmarks have been proposed to automatically evaluate their behavior.
Existing benchmarks evaluate various aspects of full-duplex dialogue, including turn-taking~\cite{Arora2024-lh,Guan-Ting2025-hx,Lin2025-jx}, response naturalness~\cite{Lin2025-ey}, and instruction following~\cite{Lin2025-ey,Zhang2026-ca}.
These studies provide important tools for evaluating full-duplex behavior.
A common approach is single-turn evaluation, in which each model responds to an isolated user input.
Such evaluation supports controlled comparison because all models receive the same local input, but it cannot reveal whether model behavior changes with preceding dialogue history.
Recent work has also extended evaluation to multi-turn settings~\cite{Lin2025-ey,Zhang2026-ca}.
However, comparable evaluation in multi-turn conversations is not straightforward and introduces additional challenges.
Dynamic evaluation with an automated examiner~\cite{Lin2025-ey} allows the conversation to adapt to each model, but the resulting dialogue trajectories differ across systems, making fine-grained comparisons between models or between humans and models difficult.
Static evaluation uses fixed user inputs to improve comparability, but these inputs may become inconsistent with the model-generated system history.
This context mismatch can make the dialogue history unnatural and affect the validity of the evaluation~\cite{Zhang2026-ca}.
% This context mismatch can affect subsequent responses and complicate the interpretation of evaluation results~\cite{Zhang2026-ca}.
% A useful multi-turn benchmark should therefore support free-running evaluation under fixed user inputs while also enabling controlled analysis of sensitivity to dialogue history.

Language and domain coverage provide additional dimensions for evaluating full-duplex behavior.
% Previous work has considered several dialogue settings, including casual conversation~\cite{Guan-Ting2025-hx,Zhang2026-ca} and task-oriented interactions such as question answering, mental-health dialogue, and reservations~\cite{Lin2025-ey,Zhang2026-ca}.
% However, cross-domain comparisons of full-duplex behavior remain limited.
% Appropriate interaction behavior can vary greatly depending on the type of conversation.
% For example, casual conversations and question answering may have different turn-taking patterns and user expectations.
% Language coverage is also limited, and most existing FDSDS benchmarks are designed mainly for English dialogues.
% %, although models for non-English languages are also being studied~\cite{Ohashi2025-ye,Abe2026-id,Zhao2026-fd}.
% Although non-English full-duplex models have begun to emerge~\cite{Ohashi2025-ye,Abe2026-id,Zhao2026-fd}, systematic evaluation resources for languages such as Japanese remain limited.
Existing benchmarks cover casual conversation~\cite{Guan-Ting2025-hx,Zhang2026-ca}, as well as question answering, mental-health dialogue, and reservation tasks~\cite{Lin2025-ey,Zhang2026-ca}.
Conversational goals and user expectations can differ across these settings, yet systematic comparisons of full-duplex behavior across domains remain limited.
Language coverage is also limited, with most existing FDSDS benchmarks focusing primarily on English.
Although non-English full-duplex models have begun to emerge~\cite{Ohashi2025-ye,Abe2026-id,Zhao2026-fd}, evaluation resources for languages such as Japanese remain limited.

In this work, we propose M3-DuplexBench, a multi-turn, multilingual, multidomain benchmark for FDSDSs.\footnote{Examples are available here: \url{https://anonymous.4open.science/w/m3-duplexbench-demo-1328/}}
% It supports English and Japanese, covers casual conversation and multi-turn question answering, and evaluates full-duplex behavior across timing and content metrics.
% A key feature of M3-DuplexBench is controlled multi-turn evaluation using teacher-forced inference, which enables fine-grained model comparison under the same coherent dialogue context.
The benchmark jointly evaluates interaction timing and response content in English and Japanese across casual conversation (Chat) and multi-turn question answering (MT-QA).
% For the question-answering domain, the English and Japanese evaluation sets are constructed from the same underlying conversations, enabling cross-lingual comparison.
The English and Japanese MT-QA data are derived from a common source corpus, enabling cross-lingual comparison.
% The benchmark supports a free-running multi-turn setting, in which each model generates its own preceding system history, as well as a reference-conditioned multi-turn setting used as an optional diagnostic for analyzing sensitivity to dialogue history and trajectory mismatch.
% Together with a single-turn setting, these conditions enable us to distinguish the effect of providing dialogue history from the effect of the particular dialogue trajectory represented by that history.
The benchmark supports three context conditions: \textit{Single}, which provides only the current user input; \textit{Multi (gen)}, which uses fixed user inputs and model-generated system history; and \textit{Multi (ref)}, which uses teacher-forced reference history.
Together, these conditions allow us to examine sensitivity to the presence of dialogue history and to the use of generated or reference system histories.
Our experiments revealed several findings:
\begin{itemize}
    \vspace{-1mm}
    \item Models exhibited distinct patterns in turn-taking, pause handling, overlap handling, and response quality.
    \item Cross-domain and cross-lingual comparisons revealed clear behavioral differences, including lower response quality and weaker separation between turn changes and user-held pauses for the evaluated Japanese models.
    \item Dialogue context affected turn-taking in a model-dependent manner, with some models showing larger changes in response timing and floor-taking behavior than others.
    \vspace{-1mm}
\end{itemize}

% \begin{table}[t]
% \centering
% \caption{Multi-turn evaluation protocols used in existing FDSDS benchmarks.
% FDB and MTR denote Full-Duplex-Bench and MTR-DuplexBench, respectively.
% $\dagger$ Full-context conditioning is applied only to dialogue quality evaluation.}
% \vspace{-2mm}
% \label{tab:benchmark_comparison}
% % \small
% \footnotesize
% \setlength{\tabcolsep}{3pt}
% \renewcommand{\arraystretch}{0.9}
% % \begin{tabular}{l@{ }c@{ }c@{ }c@{ }c@{ }c@{ }}
% \begin{tabular}{lccccc}
% \toprule
% \textbf{Multi-turn} & \textbf{FDB} & \textbf{Talking} & \textbf{FD-Bench} &  \textbf{MTR} & \textbf{Ours}\\
% \textbf{evaluation} & \cite{Guan-Ting2025-hx, Lin2025-jx, Lin2025-ey} &  \textbf{Turns}\cite{Arora2024-lh} & \cite{Peng2025-cw}&\cite{Zhang2026-ca} & \\
% \midrule
% \textit{None} (Single-turn) & \checkmark \footnotesize{v1, 1.5}& & & & \checkmark \\ \midrule
% \textit{Dynamic evaluation} & \checkmark  \footnotesize{v2}& \checkmark  \\ \midrule
% \textit{Static evaluation} \\
% ~~User context & & &\checkmark& \checkmark & \checkmark \\
% ~~Full context &  & & &\checkmark$^\dagger$& \checkmark\\
% \bottomrule
% \end{tabular}
% \vspace{-4mm}
% \end{table}

\section{RELATED WORK}
Benchmarks for FDSDSs have recently been proposed to evaluate different aspects of full-duplex interaction.
Full-Duplex-Bench evaluates smooth turn-taking, pause handling, backchanneling, and user interruption (i.e. barge-in) handling~\cite{Guan-Ting2025-hx}.
Full-Duplex-Bench v1.5 extends the evaluation to overlapping speech scenarios, including user backchannels, barge-ins, side conversations, and background speech~\cite{Lin2025-jx}.
% Several studies have also evaluated multi-turn interactions through conversations with humans~\cite{Arora2024-lh} or automated examiners~\cite{Lin2025-ey}, or using pre-collected spoken dialogue data~\cite{Peng2025-cw,Zhang2026-ca}.
Talking Turns evaluates whether spoken dialogue models can understand, predict, and perform turn-taking events using a supervised turn-taking model trained on human-human conversations as an evaluator~\cite{Arora2024-lh}.
FD-Bench evaluates FDSDSs under simulated conversations, focusing on interruption handling, response delay, and robustness to challenging acoustic conditions~\cite{Peng2025-cw}.

\subsection{MULTI-TURN EVALUATION}
Recent studies have extended spoken dialogue evaluation to multi-turn conversations.
\textit{Dynamic evaluation} allows a human user or an automated examiner to interact with the target system in real time. 
For example, Full-Duplex-Bench v2 uses an automated examiner to evaluate multi-turn interaction~\cite{Lin2025-ey}.
This enables adaptive evaluation without context mismatch described below, but different systems may receive different dialogue histories, making fine-grained comparisons across models or between humans and models difficult.
\textit{Static evaluation} instead uses pre-collected dialogue data to provide comparable user inputs across systems.
MTR-DuplexBench uses different static protocols for dialogue-quality and timing evaluation~\cite{Zhang2026-ca}.
For dialogue-quality evaluation, it uses teacher-forced inference to condition the model on reference system history, avoiding context mismatch between fixed user inputs and model-generated system history.
In contrast, its timing evaluation, including smooth turn-taking and pause handling, uses pre-collected user-side speech while allowing the model to generate its own responses.
This choice may reflect an assumption that timing decisions rely mainly on local cues and are therefore relatively insensitive to context mismatch.
However, context mismatch may also affect timing behavior in natural multi-turn interaction, a possibility not directly assessed by these protocols.
In this work, we compare timing behavior across single-turn, generated-history multi-turn, and reference-conditioned multi-turn settings for models that support reference conditioning.
% \textit{Static evaluation} instead uses pre-collected dialogue histories, providing more comparable input conditions across systems.
% MTR-DuplexBench evaluates multi-turn full-duplex conversations under such a setting~\cite{Zhang2026-ca}.
% For dialogue-quality evaluation, it uses both user- and system-side reference histories, by applying teacher-forced inference to condition the model on reference system context.
% In contrast, its timing evaluation, including smooth turn-taking and pause handling, uses only user-side dialogue history while allowing the model to generate its responses freely.
% This setting can introduce a context mismatch between the fixed user history and the model-generated system history, potentially affecting subsequent model behavior.
% Moreover, multi-turn dialogues used for timing evaluation are constructed by concatenating single-turn utterance pairs.
% % These settings are based on the assumption that timing behavior is independent of conversational context.
% These settings are based on the assumption that timing behavior is based on local decision and is independent of conversational history.
% However, whether this assumption holds in natural multi-turn interaction remains unclear.
% M3-DuplexBench instead evaluates target interaction events embedded in continuous multi-turn dialogues.
% In addition to free-running multi-turn evaluation, we compare single-turn and reference-conditioned multi-turn settings to separately examine sensitivity to the presence of dialogue history and to the particular dialogue trajectory represented by that history.

Other benchmarks focus on semantic capabilities in multi-turn settings.
For example, AudioMC evaluates long spoken conversations by focusing on capabilities such as inference memory, instruction retention, self-coherence, and voice editing~\cite{gosai2026audio}.
They complement full-duplex benchmarks, but do not directly evaluate fine-grained interaction timing such as turn shifts, pauses, backchannels, and barge-ins.

\subsection{MULTILINGUAL AND CROSS-DOMAIN EVALUATION}
Most existing full-duplex dialogue benchmarks focus primarily on English.
Recently, FDSDSs have also been developed and evaluated in other languages.
Yan et al.~\cite{Yan2026-zx} extended Full-Duplex-Bench to Chinese.
ICASSP 2026 HumDial Challenge includes both Chinese and English data~\cite{Zhao2026-fd}.
However, to our knowledge, no existing benchmark supports automatic evaluation of Japanese FDSDSs.
Moreover, cross-lingual comparisons of full-duplex timing and response behavior remain limited.

Dialogue domain is an important factor in spoken dialogue evaluation because interaction patterns and response requirements can vary across conversational settings.
Full-Duplex-Bench v1/v1.5 evaluates different interaction behaviors using different data sources.
MTR-DuplexBench also uses different data for timing and content evaluation.
Dialogue quality is evaluated on natural conversational data, whereas timing-related behaviors are evaluated on separately constructed multi-turn contexts.
Consequently, these benchmarks do not directly compare the same timing and content dimensions across dialogue domains.

Some recent studies have examined variation across domains or interaction settings.
Full-Duplex-Bench v2 covers several tasks, including Daily, Correction, Entity Tracking, and Safety, and reports turn-taking fluency and multi-turn instruction following for each task.
TurnBench evaluates end-of-turn and interruption detection across six different interaction styles, showing that turn-taking performance can depend strongly on conversation type~\cite{jiang2026turnbenchmultidomainbenchmarkturntaking}.

M3-DuplexBench supports comparison between two contrasting dialogue settings: casual conversation (Chat) and multi-turn question answering (MT-QA).
Chat domain is relatively open-ended, with ambiguous turn transitions and optional responses, whereas MT-QA has a more explicit interaction structure in which user questions typically elicit system responses.
Evaluating the same timing and response-quality dimensions across these settings enables analysis of how model behavior changes with the dialogue domain.

\begin{table}[!t]
    \small
    \centering
    \begin{spacing}{0.75}
    \caption{Comparison of benchmarks for turn-taking evaluation
    in FDSDSs. The multi-turn columns indicate protocols used for turn-taking evaluation. Dyn.: dynamic interaction; Gen.: fixed user-side context; Ref.: reference full context; X-ling.: cross-lingual comparison; X-dom.: cross-domain/setting comparison. $\triangle$: Multiple languages are supported, but cross-lingual comparison is not a primary evaluation.}
    \label{tab:benchmark_comparison}
    \end{spacing}
    % \vspace{-1mm}
    \setlength{\tabcolsep}{2pt}
    \setlength{\aboverulesep}{1pt}
    \setlength{\belowrulesep}{1pt}
    \renewcommand{\arraystretch}{0.9}
    \begin{tabular}{lcccccc} \toprule
    & \multicolumn{3}{c}{\textbf{Multi-turn}} & \multicolumn{2}{c}{\textbf{Language}} & \multicolumn{1}{c}{\textbf{Domain}} \\
    \cmidrule(l{\tabcolsep}r{\tabcolsep}){2-4} \cmidrule(l{\tabcolsep}r{\tabcolsep}){5-6} \cmidrule(l{\tabcolsep}r{\tabcolsep}){7-7} \textbf{Benchmark} & \textbf{Dyn.} & \textbf{Gen.} & \textbf{Ref.} & \textbf{Lang.} & \textbf{X-ling.} & \textbf{X-dom.} \\ \midrule
    Full-Duplex-Bench v1~\cite{Guan-Ting2025-hx} & -- & -- & -- & En & -- & -- \\
    Full-Duplex-Bench v1.5~\cite{Lin2025-jx} & -- & -- & -- & En & -- & -- \\
    Full-Duplex-Bench v2~\cite{Lin2025-ey} & \checkmark & -- & -- & En & -- & \checkmark \\
    ICASSP 2026 HumDial~\cite{Zhao2026-fd} & -- & -- & -- & En, Zh & $\triangle$ & -- \\
    Talking Turns~\cite{Arora2024-lh} & \checkmark & -- & -- & En & -- & -- \\
    FD-Bench~\cite{Peng2025-cw} & -- & \checkmark & -- & En & -- & -- \\
    MTR-DuplexBench~\cite{Zhang2026-ca} & -- & \checkmark & -- & En & -- & -- \\
    TurnBench~\cite{jiang2026turnbenchmultidomainbenchmarkturntaking} & -- & \checkmark & -- & En & -- & \checkmark \\
    \textbf{M3-DuplexBench} & -- & \checkmark & \checkmark & En, Ja & \checkmark & \checkmark \\ \bottomrule
    \end{tabular}
    \vspace{-3mm}
\end{table}

As summarized in \cref{tab:benchmark_comparison}, existing benchmarks cover complementary aspects of turn-taking evaluation.
M3-DuplexBench complements them by combining controlled multi-turn evaluation with cross-lingual and cross-domain comparisons under a common framework.

\begin{figure}[!t]
  \centering
  % \subfloat[
  %   Events included in spoken dialogues: Turn shift ($\mathrm{SHIFT}$), long pause ($\mathrm{PAUSE}$), backchanneling ($\mathrm{BC}$), and barge-in ($\mathrm{BARGE\_IN}$).
  %   \label{fig:events}
  % ]{
  \subfloat[]{
    \begin{minipage}{0.8\linewidth}
      \centering
      % \vspace{-2mm}
      \includegraphics[width=\linewidth]{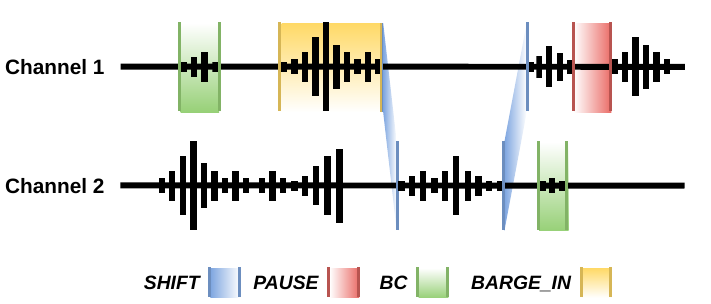}
      \vspace{-5mm}
    \end{minipage}
    \label{fig:events}
  } \\
  \vspace{-2mm}
  % \subfloat[
  %   Temporal regions used in evaluation.
  %   \label{fig:terms}
  % ]{
  \subfloat[]{
    \begin{minipage}{0.8\linewidth}
      \centering
      \includegraphics[width=\linewidth]{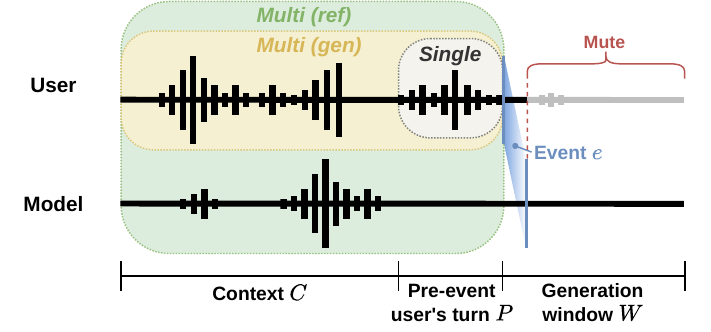}
      \vspace{-3mm}
    \end{minipage}
    \label{fig:terms}
  }
  % \caption{Overview of the proposed M3-DuplexBench evaluation framework. (a) Four interaction event types extracted from continuous spoken dialogues. (b) Evaluation protocol under three context conditions (\textit{Single}, \textit{Multi (gen)}, and \textit{Multi (ref)}).}
  \caption{Overview of the proposed M3-DuplexBench evaluation framework.
  (a) Four interaction event types extracted from continuous spoken
  dialogues: turn shift ($\mathrm{SHIFT}$), long pause ($\mathrm{PAUSE}$),
  backchanneling ($\mathrm{BC}$), and barge-in ($\mathrm{BARGE\_IN}$).
  (b) Evaluation protocol under three context conditions
  (\textit{Single}, \textit{Multi (gen)}, and \textit{Multi (ref)}).}
  \label{fig:examples}
  \vspace{-3mm}
\end{figure}

\section{M3-DUPLEXBENCH: DATA}
\label{sec:m3:data}
% M3-DuplexBench extracts evaluation events from natural and synthetic spoken dialogue datasets.
This section describes the event definitions and the datasets used in the benchmark.

\subsection{EVENT DEFINITION}
\label{subsec:m3:event}
We focus on four events: turn shift ($\mathrm{SHIFT}$), long pause ($\mathrm{PAUSE}$), backchannel ($\mathrm{BC}$), and barge-in ($\mathrm{BARGE\_IN}$), as shown in~\cref{fig:events}.
$\mathrm{TURN}$ and $\mathrm{BC}$ are inter-pausal units (IPUs) separated by silences longer than 0.5~s.
IPUs with a duration of 0.8~s or less are regarded as $\mathrm{BC}$s, while all others are considered $\mathrm{TURN}$s.
$\mathrm{PAUSE}$ denotes the interval between $\mathrm{TURN}$s produced by the same speaker.
SHIFT denotes the interval between $\mathrm{TURN}$s produced by different speakers.
For SHIFTs, overlaps of up to 0.4~s are allowed.
Speaker changes with an overlap longer than 0.4~s are regarded as $\mathrm{BARGE\_IN}$s.

Let a two-party dialogue be $D=(a^{(0)}, a^{(1)}, \mathcal{E}^{(0)}, \mathcal{E}^{(1)})$, where $a^{(c)}$ is the waveform of channel $c \in \{0,1\}$ and $\mathcal{E}^{(c)}=\{e_0^{(c)},...,e_{N^{(c)}}^{(c)}\}$ is its event sequence of length $N^{(c)}$.
An event $e \in \mathcal{E}^{(0)} \cup \mathcal{E}^{(1)}$ is written as $e=(\ell,s,t)$, where $\ell\in\{\mathrm{SHIFT},\mathrm{PAUSE},\mathrm{BC},\mathrm{BARGE\_IN}\}$ is the event type and $s,t$ are the event's start and end ti
mes.
% SHIFT events may have $t<s$ because overlaps are allowed.
We describe the procedure for evaluating models using $D$ in~\cref{sec:m3:framework}.

\subsection{DATASET}
\label{subsec:m3:dataset}
% M3-DuplexBench covers two domains, chat and multi-turn question answering (MT-QA), and two languages, English and Japanese.
We use natural spoken dialogue data for the chat domain and synthetic spoken dialogue data for the MT-QA domain.
% \cref{tab:data_statistics} summarizes the data statistics.

For the English chat domain, we use CANDOR~\cite{reece2023candor}, following MTR-DuplexBench~\cite{Zhang2026-ca}.
For the Japanese chat domain, we use MagicData, a natural speech conversation dataset of approximately 10 hours~\cite{magicdata_japanese_duplex_2025}.
%\footnote{\url{https://magichub.com/japanese}}.
Each dialogue is split into segments of approximately 120~s without splitting utterances, following~\cite{Zhang2026-ca}.

For the MT-QA domain, we use TopiOCQA, an English multi-turn QA dataset~\cite{adlakha-etal-2022-topiocqa}.
Each conversation consists of an average of 13 turns.
Unlike prior work that constructs dialogues by concatenating single-turn QA pairs~\cite{Zhang2026-ca}, we use multi-turn QA dialogues, enabling us to evaluate whether models can answer questions based on dialogue context.
% Japanese QA data was created by translating TopiOCQA.
% Therefore, the same knowledge is tested for both English and Japanese models.
We convert these text-based task dialogues into synthetic spoken dialogues using the TTS-based data generation pipeline described below.
% This pipeline can generate spoken dialogues in any language, enabling the same knowledge to be tested across multiple languages.
We create both English and Japanese data based on TopiOCQA.
The Japanese data are created by translating TopiOCQA, supporting comparison across languages.

\begin{table}[!t]
    \small
    \setlength{\tabcolsep}{2pt}
    \centering
    \setlength{\aboverulesep}{1pt}
    \setlength{\belowrulesep}{1pt}
    \renewcommand{\arraystretch}{0.9}
    \caption{Quality assessment of the evaluation data.}
    % \vspace{-1mm}
    \label{tab:audio_quality}
    % \resizebox{\linewidth}{!}{
    \begin{tabular}{lrrrrr}
        \toprule
        \textbf{Data} & \textbf{DNSMOS} & \textbf{SQ-STOI} & \textbf{SQ-PESQ} & \textbf{UTMOSv2} & \textbf{CometKiwi} \\
        \midrule
        Chat (En) & 2.44 & 0.911 & 2.282 & 2.040 & - \\
        Chat (Ja) & 3.01 & 0.912 & 2.333 & 2.557& -  \\
        MT-QA (En) & 3.17 & 0.993 & 3.499 & 2.798 & -  \\
        MT-QA (Ja) & 3.03 & 0.995 & 3.593 & 2.766 & 0.865  \\ \bottomrule
    \end{tabular}
    % }
    \vspace{-1mm}
\end{table}

\begin{table}[!t]
    \small
    \setlength{\tabcolsep}{2pt}
    \centering
    \setlength{\aboverulesep}{1pt}
    \setlength{\belowrulesep}{1pt}
    \renewcommand{\arraystretch}{0.9}
    % \caption{Data statistics of M3-DuplexBench. Values are reported as test subset / full dataset. %$\dag$ indicates statistics used in the main experiments.}
    \begin{spacing}{0.75}
    \caption{Data statistics of M3-DuplexBench. Values are reported as test / dev set. The test sets are used in the main experiments.}
    \label{tab:data_statistics}
    \end{spacing}
    % \vspace{-1mm}
    % \resizebox{\linewidth}{!}{
    \begin{tabular}{lcccc}
        \toprule
        & \textbf{Chat (En)} & \textbf{Chat (Ja)} & \textbf{MT-QA (En)} & \textbf{MT-QA (Ja)} \\ \midrule
        Speech & Real & Real & Synthetic & Synthetic \\
        Source  & CANDOR & MagicData & TopiOCQA & TopiOCQA \\
        Source type & Speech & Speech & Text & Text \\ \midrule
        \# Dialogues & ~31 / 169 & ~33 / ~125 & ~49 / 91 & ~45 / 148 \\
        Total Dur. (h) & 1.04 / 5.59 & 1.01 / 3.86 & 1.02 / 2.01 & 1.01 / 4.31 \\
        Mean Dur. (s) & 120.8 / 119.1 & 110.2 / 111.2 & 74.9 / 79.5 & 80.8 / 104.8 \\ \midrule
        \# $\mathrm{SHIFT}$ & 260 / 1100 & 349 / 1495 & 529 / 910 & 502 / 1760 \\ 
        \# $\mathrm{PAUSE}$ & 400 / 2367 & 333 / 1378 & -- / -- & -- / -- \\
        \# $\mathrm{BC}$ & 110 / ~576 & 121 / 943 & -- / -- & -- / -- \\ %\hdashline
        % \# BARGE\_IN & ~\dag\red{61} / ~384 & \dag134 / ~785 & ~190 / ~521 & ~188 / ~945 \\ \bottomrule % without cutoff point augumentation (drop3-6)
        \# $\mathrm{BARGE\_IN}$ & 119 / ~613 & 269 / 1297 & 380 / 683 & 392 / 1421 \\ \bottomrule % with cuttoff point augmentation (drop 3 & 6)
    \end{tabular}
    % }
    \vspace{-3mm}
\end{table}

We additionally assess the quality of the dialogue data using objective metrics, as summarized in \cref{tab:audio_quality}.
Following prior work on full-duplex dialogue dataset~\cite{nakata2026duplexchatconstructingspeakerseparatedfullduplex}, we report DNSMOS~\cite{9414878}, SQ-STOI, and SQ-PESQ~\cite{kumar2023torchaudio} to assess speech quality.
We additionally report UTMOSv2~\cite{baba2024utmosv2} to assess speech naturalness and COMETKiwi~\cite{rei-etal-2022-cometkiwi}, a reference-free translation quality estimation metric, for the Japanese MT-QA data.
% \red{Synthetic＞Realの解釈をいれる。1〜2文}
During data inspection, we found that a subset of the CANDOR recordings (Chat En) was particularly noisy, which lowered the overall DNSMOS score.
We therefore compute DNSMOS at the dialogue level and select approximately one hour of the highest-scoring dialogues for each domain and language as the test set.
The remaining dialogues are kept speaker-disjoint from the test set and can be used as a development set.
For MT-QA datasets, test-set selection was performed independently for each language, resulting in 23 dialogue IDs shared between the English and Japanese test sets.
\cref{tab:data_statistics} summarizes the data statistics.

\subsubsection{Synthetic dialogue generation pipeline}
Synthetic dialogue generation has recently been explored for training FDSDSs, including text-based dialogue synthesis~\cite{suresh-etal-2025-diasynth} and synthetic spoken dialogue generation~\cite{wang25x_interspeech,lee-etal-2025-behavior,Ohashi2025-ye}.
Following these works, we construct synthetic MT-QA data from source text dialogues and reference spoken dialogues through four steps.
% We used TopiOCQA for text dialogues and CANDOR and MagicData for English and Japanese reference spoken dialogues, respectively.
We use TopiOCQA as the source text dialogues, and CANDOR and MagicData as English and Japanese reference spoken dialogues, respectively.

\begin{enumerate}
    \vspace{-1mm}
    \item We analyze the reference spoken dialogues to obtain statistics of turn-transition timing for synthetic dialogue construction.
    \item We edit the text dialogues. Written dialogues are converted into spoken-style dialogues using a large language model (LLM) and translated into Japanese when needed. We use Gemma 4 31B~\cite{gemma4_31b}.
    \item We synthesize speech for each utterance using a CosyVoice2-based TTS model~\cite{du2024cosyvoice}. For speaker conditioning, we randomly select reference speech from the reference dialogues.
    \item Finally, we place the synthesized utterances on a timeline, sampling $\mathrm{SHIFT}$ timing from the statistics obtained from the reference dialogues.
    % Finally, we place the synthesized utterances on a timeline. $\mathrm{PAUSE}$, $\mathrm{SHIFT}$, and BC events are sampled from the statistics obtained from the reference dialogues. This produces synthetic spoken dialogues with timing characteristics similar to those observed in natural dialogues.
    \vspace{-1mm}
\end{enumerate}
$\mathrm{BARGE\_IN}$ events are created separately because they rarely appear in ordinary spoken dialogue data.
Following Hu et al.~\cite{hu25f_interspeech}, we create a $\mathrm{BARGE\_IN}$ sample by cutting off the system utterance at a sampled interruption point and shifting the user utterance to start from that point.
This creates a controlled overlap where the user starts speaking while the system is still speaking.
The synthetic MT-QA data are used to evaluate $\mathrm{SHIFT}$ and $\mathrm{BARGE\_IN}$ events.
$\mathrm{PAUSE}$ and $\mathrm{BC}$ are evaluated only using the natural Chat data.

\section{M3-DUPLEXBENCH: EVALUATION FRAMEWORK}
\label{sec:m3:framework}

\subsection{INFERENCE}
\label{subsec:m3:inference}
For the Chat domain, we use target events from both speaker channels (i.e., $e \in \mathcal{E}^{(0)} \cup \mathcal{E}^{(1)}$), since the two speakers alternate the user and system roles.
For the MT-QA domain, we use only events from the user channel (i.e., $e \in \mathcal{E}^{(0)}$), because the dialogue roles are fixed.
For each event $e$, we define three regions: context $C$, pre-event user's turn $P$, and generation window $W$, as shown in~\cref{fig:terms}.
The generation window starts at the event onset $s$ and has a fixed duration $\Delta$, i.e., $W=[s,s+\Delta]$.
The Pre-event user's turn $P$ is the short segment before the event, i.e., $P=[b,s]$.
For $\mathrm{SHIFT}$ and $\mathrm{PAUSE}$, $b$ is the start time of a turn immediately preceding or containing the event, respectively.
% For $\mathrm{BC}$ and $\mathrm{BARGE\_IN}$, the user is speaking during a system turn, and thus there are no pre-event user's turn, so $P=\emptyset$ and we set $b=s$.
For $\mathrm{BC}$ and $\mathrm{BARGE\_IN}$, the user event occurs during a system turn in the reference dialogue.
We therefore use no separate pre-event user turn, set $P=\emptyset$, and set $b=s$.
The context $C$ is the dialogue history before $P$.
Given a context length $L$, $C=[\tau(L),b]$, where $\tau(L)$ is the start time of the earliest turn overlapping the time window $[b-L,b]$.
In the experiment, we set $\Delta=10$~s and $L=120$~s.
We consider three context conditions:
\begin{itemize}
    \vspace{-1mm}
    \item \textbf{\textit{Single}} (single-turn): provides no dialogue context and uses only the current event input. The model receives user-side audio in $(P, W)$. This corresponds to a single-turn evaluation setting in previous works~\cite{Guan-Ting2025-hx,Lin2025-jx}. $\mathrm{BC}$ and $\mathrm{BARGE\_IN}$ are not evaluated under this condition because $P$ does not exist.
    \item \textbf{\textit{Multi (gen)}} (free-running multi-turn): provides user-side speech history as a context. The model receives user-side audio in $(C, P, W)$, while system-side audio is not provided. The model state before the target event is induced by the model's own generated responses.
    \item \textbf{\textit{Multi (ref)}} (reference-conditioned multi-turn): provides both user- and system-side speech history through teacher-forced inference. The model state is forced using both speaker channels in $(C, P)$ before receiving user-side audio in $W$.
    \vspace{-1mm}
\end{itemize}
For all conditions, user-side audio within $W$ is retained only during the target event interval and muted outside that interval.
After inference, we evaluate the speech generated by the model using word-level timestamps.
We first use Whisper ASR~\cite{radford2022robustspeechrecognitionlargescale,faster_whisper} to transcribe the generated speech, then apply Montreal Forced Aligner~\cite{mcauliffe17_interspeech} to align the ASR transcript with the generated audio and obtain timestamps.
These processes were applied individually to each channel.

\begin{table}[!t]
    \centering
    \small
    \renewcommand{\arraystretch}{0.9}
    \caption{Evaluation dimensions and metrics in M3-DuplexBench.}
    \vspace{-1mm}
    \label{tab:eval_metrics}
    \begin{tabular}{lll}
        \hline
        \textbf{Category} & \textbf{Dimension} & \textbf{Metrics} \\ \hline
        Timing & Smooth Turn-Taking & TOR $\uparrow$, Latency $\downarrow$, MAE $\downarrow$ \\
        & Pause Handling & TOR $\downarrow$, Latency $\uparrow$ \\
        & User Backchannel & Stop Latency $\uparrow$ \\
        & User Barge-in & Stop Latency $\downarrow$ \\
        \hline
        Content & Response Relevance & LLM-as-a-Judge$\uparrow$  \\
        & Context Consistency &  LLM-as-a-Judge$\uparrow$ \\
        & QA Score &  LLM-as-a-Judge$\uparrow$ \\ \hline
    \end{tabular}
    \vspace{-3mm}
\end{table}

\subsection{EVALUATION METRICS}
\label{subsec:m3:metrics}
Table~\ref{tab:eval_metrics} summarizes the evaluation metrics used in M3-DuplexBench.
% We evaluate models based on two aspects: timing and content.
\subsubsection{Timing Metrics}
For timing evaluation, each dimension corresponds to one event type: \textit{Smooth Turn-Taking} to $\mathrm{SHIFT}$, \textit{Pause Handling} to $\mathrm{PAUSE}$, \textit{User Backchannel} to $\mathrm{BC}$, and \textit{User Barge-in} to $\mathrm{BARGE\_IN}$.
We follow Full-Duplex-Bench~\cite{Guan-Ting2025-hx} and use the Takeover Rate (TOR) and latency for \textit{Smooth Turn-Taking} and \textit{Pause Handling}.
Note that our computation of TOR and latency is not exactly the same as \cite{Guan-Ting2025-hx}.
In particular, their definition can count speech in all regions prior to the event as a takeover, thereby increasing TOR and reducing latency.
% \footnote{Our computation is not exactly the same as \cite{Guan-Ting2025-hx}. For example, their definition can count speech in the pre-event region as a takeover, which tends to increase TOR and reduce latency.}.

TOR measures the fraction of samples in which the model takes the turn within the scoring interval.
For \textit{Smooth Turn-Taking}, this interval extends to the end of the $\Delta$-second generation window.
For \textit{Pause Handling}, it is limited to the actual user pause in the reference dialogue, ending when the user resumes speaking.
A response after the end of this pause is therefore not counted as a takeover, even if it occurs within the generation window.
To allow for overlap, response onsets up to 0.4~s before the event onset are also considered.
Response latency is measured from the event onset to the onset of the model response, and is computed over samples with a response onset no later than $\Delta$~s after the event onset.
Samples without a response by this time are excluded from the latency calculation.
For \textit{Smooth Turn-Taking}, a higher TOR and a lower latency indicate more frequent and faster responses to turn transitions.
For \textit{Pause Handling}, a lower TOR indicates fewer responses during user-held pauses, whereas a higher latency indicates a longer delay before responding among samples with an observed response.
For \textit{Smooth Turn-Taking}, we additionally report the mean absolute error (MAE) of the model's response latency relative to the event-specific human reference latency.
Unlike mean latency, MAE measures how closely its response timing matches the event-specific human timing.

For \textit{User Backchannel} and \textit{User Barge-in}, we follow Full-Duplex-Bench v1.5~\cite{Lin2025-jx} and measure stop latency, defined as the time from the event onset to the point where the model stops speaking.
Stop latency is computed only for samples in which the model is speaking at the event onset.
Samples in which the model is not speaking are excluded rather than assigned a zero latency.
For backchannels, a higher stop latency is better because the model should continue speaking.
For barge-ins, a lower stop latency is better because the model should stop quickly after the user interruption.
% Under \textit{Multi (gen)}, the generated system speech may differ in timing from the reference dialogue. Consequently, the set of samples eligible for stop-latency evaluation can differ across models, and the reported means characterize stopping behavior conditional on the model being active at the event onset. Under \textit{Multi (ref)}, reference conditioning ensures an ongoing system utterance at the event onset. However, continuing generation from a partially supplied reference utterance can sometimes produce unnatural speech, which should be considered when interpreting results under this condition.

\begin{table*}[!t]
    \small
    \begin{spacing}{0.75}
    \caption{Multi-turn evaluation results. Values are reported as mean$\pm$95\% confidence interval using Student's t-distribution. Smooth turn-taking MAE is computed over responded events  against the event-specific reference latency. Stop latency is computed only for events at which the model is speaking. $\dag$ indicates estimates based on fewer than 20 eligible events. \textbf{Bold}: best value within each language, excluding stop-latency estimates marked with $\dag$. Cell shading indicates relative performance within each metric and language (darker color indicates better performance).}
    \label{tab:multi_turn}
    \end{spacing}
    \setlength{\tabcolsep}{2.2pt}
    \setlength{\aboverulesep}{1pt}
    \setlength{\belowrulesep}{1pt}
    \centering
    \renewcommand{\arraystretch}{0.9}
    \begin{tabular}{llccccccc@{}p{6pt}@{}ccc@{}p{6pt}@{}ccc}
        \toprule
        & \multirow{6}{*}{\textbf{Model}} & \multicolumn{7}{c}{\textbf{Chat (Real data)}} & & \multicolumn{6}{c}{\textbf{Multi-turn QA (Synthetic data)}} \\
        \cmidrule(l{\tabcolsep}r{\tabcolsep}){3-9} \cmidrule(l{\tabcolsep}r{\tabcolsep}){11-17}
        & & \multicolumn{3}{c}{\textit{Smooth}} & \multicolumn{2}{c}{\textit{Pause}} & \multicolumn{1}{c}{\textit{User Back-}} & \multicolumn{1}{c}{\textit{User}} & & \multicolumn{2}{c}{\textit{Smooth}} & \multicolumn{1}{c}{\textit{User}} & & \multicolumn{3}{c}{\multirow{2}{*}{\textit{Response Quality}}} \\[-2pt]
        & & \multicolumn{3}{c}{\textit{Turn-Taking}} & \multicolumn{2}{c}{\textit{Handling}} & \multicolumn{1}{c}{\textit{channel}} & \multicolumn{1}{c}{\textit{Barge-in}} & & \multicolumn{2}{c}{\textit{Turn-Taking}} & \multicolumn{1}{c}{\textit{Barge-in}} & & & & \\
        \cmidrule(l{\tabcolsep}r{\tabcolsep}){3-5} \cmidrule(l{\tabcolsep}r{\tabcolsep}){6-7} \cmidrule(l{\tabcolsep}r{\tabcolsep}){8-8} \cmidrule(l{\tabcolsep}r{\tabcolsep}){9-9} \cmidrule(l{\tabcolsep}r{\tabcolsep}){11-12} \cmidrule(l{\tabcolsep}r{\tabcolsep}){13-13} \cmidrule(l{\tabcolsep}r{\tabcolsep}){15-17}
        & & \multirow{2}{*}{TOR$\uparrow$} & \multirow{2}{*}{Latency$\downarrow$} & \multirow{2}{*}{MAE$\downarrow$} & \multirow{2}{*}{TOR$\downarrow$} & \multirow{2}{*}{Latency$\uparrow$} & \multicolumn{1}{c}{Stop} & \multicolumn{1}{c}{Stop} & & \multirow{2}{*}{TOR$\uparrow$} & \multirow{2}{*}{Latency$\downarrow$} & \multicolumn{1}{c}{Stop} & & \multirow{2}{*}{Rel.$\uparrow$} & \multirow{2}{*}{Cons.$\uparrow$} & \multirow{2}{*}{QA$\uparrow$} \\[-2pt]
        & & & & & & & \multicolumn{1}{c}{Latency$\uparrow$} & \multicolumn{1}{c}{Latency$\downarrow$} & & & & \multicolumn{1}{c}{Latency$\downarrow$} & & & & \\
        \midrule
        \multirow{7}{*}{\rotatebox{90}{\textbf{En}~}} & DuplexCascade & \heat{0}{69.6{\tiny$\pm$5.6}} & \heat{0}{4.92{\tiny$\pm$0.43}} & \heat{0}{4.00{\tiny$\pm$0.42}} & \heat{40}{\textbf{0.3}{\tiny$\pm$0.5}} & \heat{40}{\textbf{7.58}{\tiny$\pm$0.28}} & \gray{0.41{\tiny$\pm$0.25}}{\tiny}$\dag$ & \gray{0.35{\tiny$\pm$0.21}}{\tiny}$\dag$ && \heat{17}{85.8{\tiny$\pm$3.0}} & \heat{0}{2.99{\tiny$\pm$0.17}} & \heat{0}{2.91{\tiny$\pm$0.41}} & & \heat{10}{61.3{\tiny$\pm$2.2}} & \heat{22}{60.2{\tiny$\pm$2.5}} & \heat{31}{24.2{\tiny$\pm$2.9}} \\
        & Freeze-Omni & \heat{16}{80.0{\tiny$\pm$4.9}} & \heat{21}{2.66{\tiny$\pm$0.30}} & \heat{28}{2.21{\tiny$\pm$0.27}} & \heat{0}{61.8{\tiny$\pm$4.8}} & \heat{2}{1.90{\tiny$\pm$0.48}} & \heat{13}{2.21{\tiny$\pm$0.45}} & \heat{14}{1.71{\tiny$\pm$0.39}} && \heat{15}{85.4{\tiny$\pm$3.0}} & \heat{11}{2.24{\tiny$\pm$0.19}} & \heat{12}{2.43{\tiny$\pm$0.26}} & & \heat{0}{56.0{\tiny$\pm$2.3}} & \heat{5}{53.5{\tiny$\pm$2.6}} & \heat{0}{17.0{\tiny$\pm$2.7}} \\
        & BayLing-Duplex & \heat{29}{88.8{\tiny$\pm$3.9}} & \heat{30}{1.67{\tiny$\pm$0.14}} & \heat{40}{\textbf{1.42}{\tiny$\pm$0.19}} & \heat{19}{33.3{\tiny$\pm$4.6}} & \heat{3}{2.05{\tiny$\pm$0.27}} & \heat{6}{1.68{\tiny$\pm$0.59}} & \heat{14}{1.69{\tiny$\pm$0.58}} && \heat{33}{91.7{\tiny$\pm$2.4}} & \heat{16}{1.89{\tiny$\pm$0.12}} & \heat{28}{1.76{\tiny$\pm$0.19}} & & \heat{11}{61.9{\tiny$\pm$2.2}} & \heat{0}{51.4{\tiny$\pm$2.9}} & \heat{2}{17.5{\tiny$\pm$2.6}} \\
        & Moshi & \heat{5}{73.1{\tiny$\pm$5.4}} & \heat{31}{1.60{\tiny$\pm$0.35}} & \heat{29}{2.13{\tiny$\pm$0.31}} & \heat{24}{25.5{\tiny$\pm$4.3}} & \heat{22}{4.82{\tiny$\pm$0.42}} & \heat{8}{1.83{\tiny$\pm$0.63}} & \heat{40}{\textbf{1.03}{\tiny$\pm$0.22}} && \heat{0}{79.8{\tiny$\pm$3.4}} & \heat{18}{1.79{\tiny$\pm$0.25}} & \heat{16}{2.26{\tiny$\pm$0.27}} & & \heat{30}{72.3{\tiny$\pm$2.3}} & \heat{31}{64.0{\tiny$\pm$2.8}} & \heat{20}{21.6{\tiny$\pm$2.8}} \\
        & Moshi-RL & \heat{40}{\textbf{95.8}{\tiny$\pm$2.5}} & \heat{40}{\textbf{0.58}{\tiny$\pm$0.15}} & \heat{39}{1.47{\tiny$\pm$0.20}} & \heat{24}{25.5{\tiny$\pm$4.3}} & \heat{4}{2.22{\tiny$\pm$0.24}} & \heat{0}{1.25{\tiny$\pm$0.35}} & \heat{37}{1.12{\tiny$\pm$0.22}} && \heat{40}{\textbf{94.3}{\tiny$\pm$2.0}} & \heat{30}{1.02{\tiny$\pm$0.18}} & \heat{24}{1.95{\tiny$\pm$0.24}} & &  \heat{40}{\textbf{77.9}{\tiny$\pm$2.3}} & \heat{34}{65.1{\tiny$\pm$2.7}} & \heat{30}{24.0{\tiny$\pm$2.8}} \\
        & PersonaPlex & \heat{30}{89.2{\tiny$\pm$3.8}} & \heat{32}{1.41{\tiny$\pm$0.25}} & \heat{36}{1.69{\tiny$\pm$0.24}} & \heat{19}{33.3{\tiny$\pm$4.6}} & \heat{9}{3.00{\tiny$\pm$0.38}} & \heat{38}{4.07{\tiny$\pm$0.85}} & \heat{0}{2.06{\tiny$\pm$0.63}} && \heat{29}{90.2{\tiny$\pm$2.5}} & \heat{40}{0.35{\tiny$\pm$0.02}} & \heat{40}{\textbf{1.28}{\tiny$\pm$0.16}} & & \heat{37}{76.5{\tiny$\pm$2.3}} & \heat{40}{\textbf{67.6}{\tiny$\pm$2.7}} & \heat{37}{25.5{\tiny$\pm$2.7}} \\
        & PersonaPlex-RL & \heat{36}{93.1{\tiny$\pm$3.1}} & \heat{36}{0.97{\tiny$\pm$0.24}} & \heat{37}{1.64{\tiny$\pm$0.24}} & \heat{16}{36.5{\tiny$\pm$4.7}} & \heat{0}{1.61{\tiny$\pm$0.20}} & \heat{40}{\textbf{4.19}{\tiny$\pm$0.89}} & \heat{8}{1.86{\tiny$\pm$0.54}} && \heat{30}{90.7{\tiny$\pm$2.5}} & \heat{40}{\textbf{0.32}{\tiny$\pm$0.05}} & \heat{35}{1.50{\tiny$\pm$0.20}} & & \heat{40}{77.7{\tiny$\pm$2.3}} & \heat{38}{66.8{\tiny$\pm$2.7}} & \heat{40}{\textbf{26.2}{\tiny$\pm$2.8}} \\
        \midrule\midrule
        \multirow{2}{*}{\rotatebox{90}{\textbf{Ja}~}} & J-Moshi & \heat{0}{31.5{\tiny$\pm$4.9}} & \heat{40}{\textbf{1.90}{\tiny$\pm$0.47}} & \heat{40}{\textbf{1.93}{\tiny$\pm$0.42}} & \heat{40}{\textbf{22.8}{\tiny$\pm$4.5}} & \heat{40}{\textbf{7.02}{\tiny$\pm$0.50}} & \heat{0}{2.51{\tiny$\pm$1.23}} & \heat{40}{\textbf{2.35}{\tiny$\pm$0.52}} && \heat{0}{77.9{\tiny$\pm$3.6}} & \heat{40}{\textbf{0.87}{\tiny$\pm$0.15}} & \heat{40}{\textbf{2.04}{\tiny$\pm$0.27}} & & \heat{40}{\textbf{49.7}{\tiny$\pm$2.0}} & \heat{40}{\textbf{63.9}{\tiny$\pm$2.5}} & \heat{40}{\textbf{11.8}{\tiny$\pm$2.2}} \\
        & LLM-jp-Moshi & \heat{40}{\textbf{84.0}{\tiny$\pm$3.9}} & \heat{0}{2.26{\tiny$\pm$0.29}} & \heat{0}{2.11{\tiny$\pm$0.25}} & \heat{0}{69.1{\tiny$\pm$5.0}} & \heat{0}{0.87{\tiny$\pm$0.36}} & \heat{40}{\textbf{3.07}{\tiny$\pm$0.64}} & \heat{0}{2.73{\tiny$\pm$0.37}} && \heat{40}{\textbf{84.1}{\tiny$\pm$3.2}} & \heat{0}{1.87{\tiny$\pm$0.22}} & \heat{0}{2.33{\tiny$\pm$0.28}} & & \heat{0}{45.4{\tiny$\pm$2.0}} & \heat{0}{57.5{\tiny$\pm$2.3}} & \heat{0}{10.5{\tiny$\pm$2.0}} \\
        \bottomrule
    \end{tabular}
    \vspace{-3mm}
\end{table*}

\subsubsection{Content metrics}
% For content evaluation, we evaluate model responses to $\mathrm{SHIFT}$ events.
% Several studies have used LLMs to evaluate FDSDS responses~\cite{Guan-Ting2025-hx,Zhang2026-ca}.
% We measure response relevance, context consistency, and QA score on a 0--2 scale, using LLM-as-a-Judge with OpenAI GPT-5 nano~\cite{openai_gpt5_nano}.
% %\footnote{\url{https://developers.openai.com/api/docs/models/gpt-5-nano}}
% % The LLM-as-a-Judge scores are originally rated on a 0–2 scale and then normalized to the range [0, 1] for reporting.
% There scores are normalized to a 0--100 scale for reporting.
% % These scores are normalized to the range [0, 1] for reporting.
% Response relevance judges whether the model response addresses the current user utterance.
% Context consistency judges whether the response is consistent with the previous dialogue context.
% QA score judges whether the response matches the reference answer in task-oriented samples.
For content evaluation, we assess model responses to $\mathrm{SHIFT}$ events using GPT-5 nano~\cite{openai_gpt5_nano} ($\texttt{gpt-5-nano-2025-08-07}$).
Each response is rated on a 0–2 scale along three dimensions.
\textit{Response relevance} uses the immediately preceding user utterance and the model response. Ratings indicate an irrelevant or non-meaningful response (0), a partially relevant, vague, generic, or incomplete response (1), or a relevant and responsive answer (2), without assessing factual correctness.
\textit{QA score} additionally uses the reference answer.
Ratings indicate an incorrect, unsupported, or non-answer response (0), a partially correct or incomplete answer (1), or a correct or semantically equivalent answer (2).
\textit{Context consistency} uses the preceding reference dialogue history and the model response.
Ratings indicate contradiction with the supplied context (0), no contradiction but insufficient contextual grounding (1), or consistency with contextual grounding (2).
% This metric assesses agreement with the reference history, rather than consistency with the model's own generated history.
These scores are normalized to a 0--100 scale for reporting.

\section{EXPERIMENTS}
\label{sec:exp}
\subsection{MODELS}
We evaluate open-source FDSDSs in English and Japanese.
% We focus on Moshi-based models~\cite{Defossez2024-zr} because \textit{Full} setting requires direct conditioning on parallel user and system speech streams.
% Extending the benchmark to other model architectures, including cascaded systems such as Freeze-Omni~\cite{Wang2024-hr}, is left for future work.
% For English, we evaluate Moshi and PersonaPlex~\cite{Roy2026-db}.
% For Japanese, we evaluate J-Moshi~\cite{Ohashi2025-ye} and LLM-jp-Moshi~\cite{Abe2026-id}.
For English, we evaluate Freeze-Omni~\cite{Wang2024-hr}, DuplexCascade~\cite{yang2026duplexcascade},
Moshi~\cite{Defossez2024-zr}, PersonaPlex~\cite{Roy2026-db}, Moshi-RL, PersonaPlex-RL~\cite{ohashi2026multi},
% MoshiRAG~\cite{chien2026moshirag},
and BayLing-Duplex~\cite{fang2026bayling}.
Moshi is an end-to-end full-duplex model based on a 7B text-based LLM Helium, and a neural audio codec Mimi.
It models parallel streams of user speech tokens, system speech tokens, and system text tokens. %\footnote{Since Moshi is trained to start with greetings, it often overlaps with the user's first utterance. Therefore, in \textit{None} and \textit{User} settings, 5 seconds of silence is prepended to the user's speech.}
PersonaPlex is based on the Moshi architecture and supports speaker and role control through voice and text prompts.
% Freeze-Omni is a cascaded full-duplex model based on voice activity detection (VAD), streaming speech encoders and decoders, and a frozen text LLM.
Moshi-RL and PersonaPlex-RL further trained these models with reinforcement learning to improve multiple aspects of conversational interactivity.
% MoshiRAG augments Moshi with asynchronous knowledge retrieval for improving response factuality.
BayLing-Duplex models user speech, system text, and system speech within a single autoregressive model.
Freeze-Omni is a cascaded full-duplex model based on voice activity detection (VAD), streaming speech encoders and decoders, and a frozen text LLM.
DuplexCascade uses a streaming cascaded ASR--LLM--TTS architecture with LLM-based turn-taking control.
J-Moshi and LLM-jp-Moshi are Moshi-based models fine-tuned on different Japanese spoken dialogue datasets.

\subsection{RESULTS}
\subsubsection{Overall Benchmark Performance}
\cref{tab:multi_turn} summarizes the main results under the \textit{Multi (gen)} setting.
Among the English models, Moshi-RL showed the highest smooth turn-taking TOR and the lowest mean latency in Chat.
PersonaPlex and PersonaPlex-RL also achieved high TORs with relatively short latencies.
DuplexCascade was more conservative, showing the lowest pause-handling TOR and the longest pause-handling latency, but also the slowest turn-taking responses.
Response speed did not always correspond to closer agreement with human timing: BayLing-Duplex obtained the lowest MAE in English Chat (1.42~s), closely followed by Moshi-RL (1.47~s).
% Thus, responding earlier did not necessarily imply closer alignment with the event-specific human response timing.

Overlap handling also varied across models.
In Chat, PersonaPlex and PersonaPlex-RL showed relatively large differences between stop latencies of user backchannels and barge-ins, continuing longer during backchannels while stopping sooner for barge-ins.
This pattern suggests a stronger ability to distinguish listener feedback from interruptions.
In contrast, Moshi and Moshi-RL had much shorter stop latencies of user backchannels, indicating a greater tendency to stop in response to user speech, even when it served as listener feedback.

For response quality, PersonaPlex, PersonaPlex-RL, and Moshi-RL achieved relatively high relevance scores, while consistency scores among the English Moshi-family models were broadly comparable.
However, QA score remained below 30 even for the best-performing model.
This is partly attributable to the difficulty of TopiOCQA, which contains many knowledge-intensive questions originally designed for open-domain information retrieval.
Indeed, the original TopiOCQA study reported a best F1 score of 31.8\% for text-based LLMs without external retrieval~\cite{adlakha-etal-2022-topiocqa}.
Freeze-Omni, DuplexCascade, and BayLing-Duplex generally obtained lower response quality scores than Moshi-based models.
These differences do not necessarily reflect content quality alone: delayed or long responses were sometimes truncated by the fixed generation window.
Responses stating an inability to answer were also observed and could receive low QA scores despite being relevant and consistent.
% These lower scores, however, do not necessarily indicate poorer response content alone.
% These models tended to respond relatively late or generate long responses, causing part of the answer to fall outside the fixed generation window.
% Meta-level responses (e.g., ``Since I don't have any information from 2019, I can't answer that question.") were also observed.
% Such behaviors can reduce response quality scores even when the generated response is relevant and consistent.

% The two Japanese models also showed contrasting interaction characteristics.
% LLM-jp-Moshi achieved higher smooth turn-taking TORs than J-Moshi, whereas J-Moshi showed better pause handling and lower smooth turn-taking latency.
% J-Moshi also achieved higher response relevance, context consistency, and QA accuracy in MT-QA.
% Domain- and language-dependent differences are examined in the following subsection.
The Japanese models showed contrasting interaction characteristics.
LLM-jp-Moshi achieved higher smooth turn-taking TORs, whereas J-Moshi showed better pause handling, lower mean turn-taking and barge-in stop latencies, and higher MT-QA response-quality scores.

\begin{figure}[!t]
  \centering
  \subfloat[]{
    \begin{minipage}{1\linewidth}
      \centering
      \includegraphics[width=\linewidth]{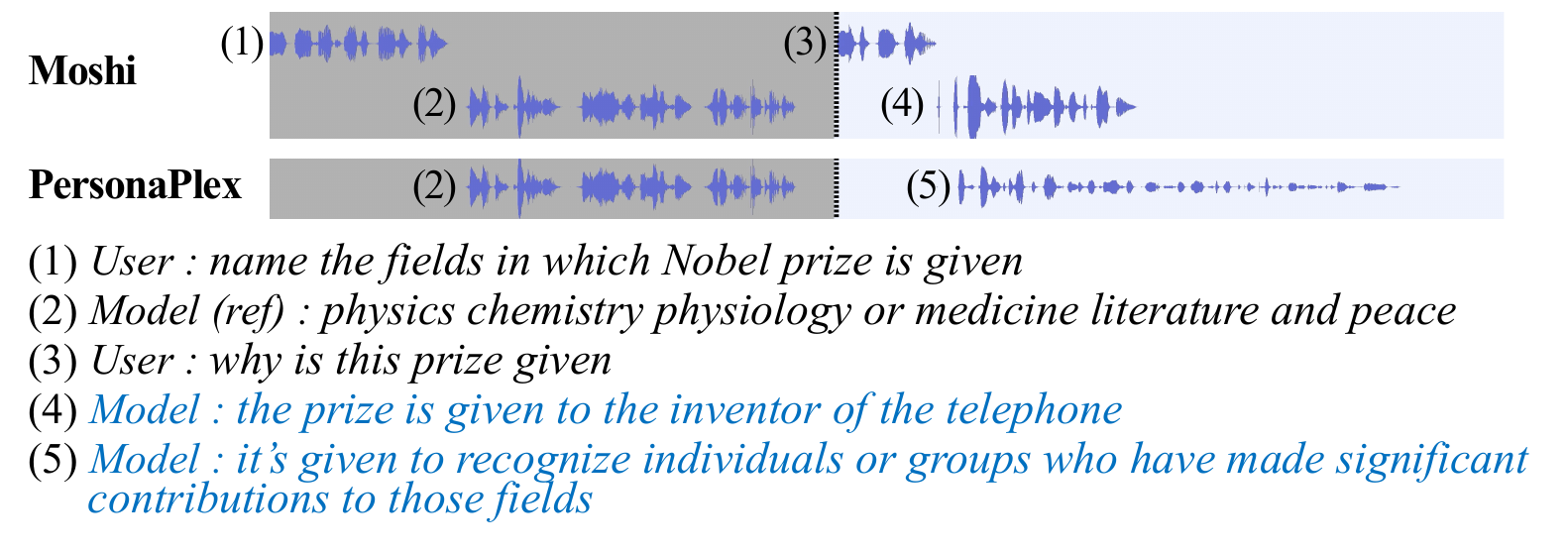}
      \vspace{-7mm}
    \end{minipage}
    \label{fig:case_study_model_en}
  } \\
  \vspace{-2mm}
  \subfloat[]{
    \begin{minipage}{1\linewidth}
      \centering
      \includegraphics[width=\linewidth]{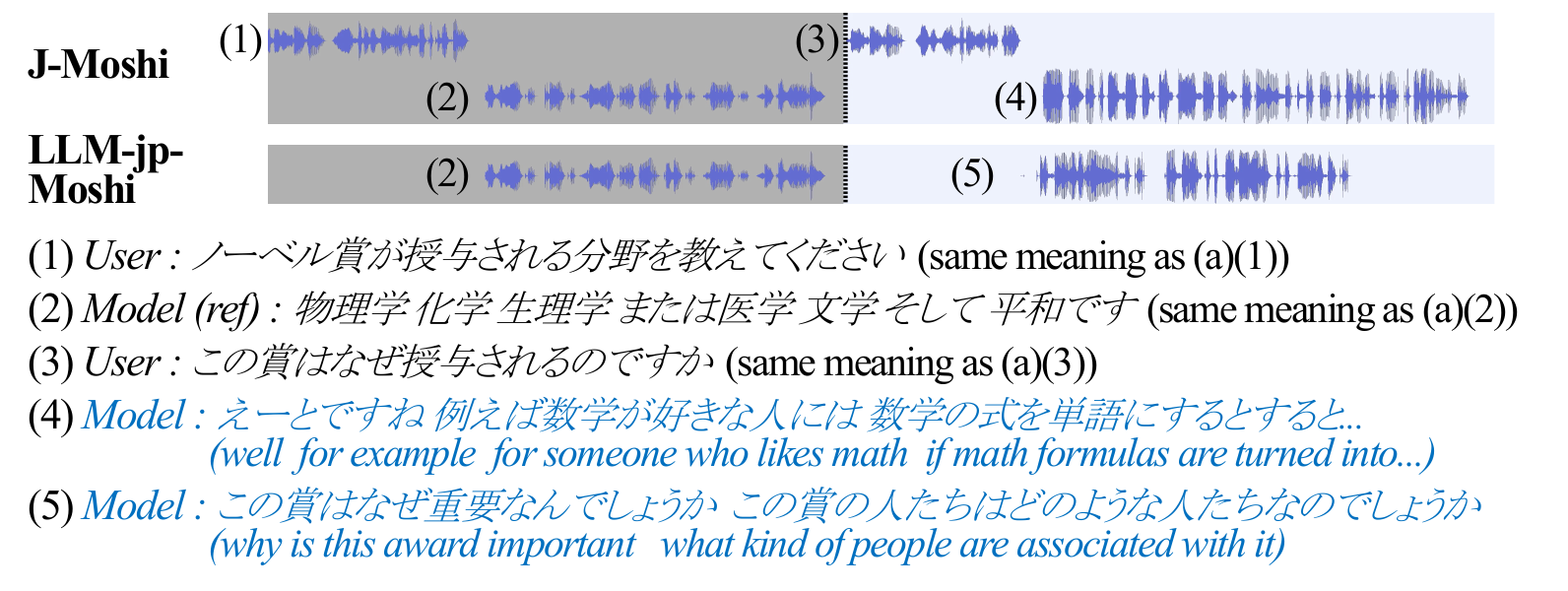}
      \vspace{-7mm}
    \end{minipage}
    \label{fig:case_study_model_ja}
  }
  \caption{Case studies of model responses in MT-QA under \textit{Multi (ref)} setting. (a) English examples. (b) Japanese examples. The gray region denotes the teacher-forced reference context. In these examples, English models generated responses relevant to the current question, whereas Japanese models produced less relevant responses, showing the cross-lingual gap in context-dependent answer generation.}
  \label{fig:case_study_model}
  \vspace{-1mm}
\end{figure}

\subsubsection{Cross-Domain and Cross-Lingual Differences}
\label{subsub:cross_domain_lingual}
% We examined differences across dialogue domains and languages.
% For cross-domain comparison, we focus on smooth turn-taking,
% which is evaluated in both chat and MT-QA.
% Overall, the models tended to achieve higher TOR in MT-QA than in chat.
% For example, DuplexCascade increased its TOR from 69.6\% in Chat to 85.8\% in MT-QA while reducing its mean latency from 4.92 to 2.99~s, and J-Moshi increased its TOR from 31.5\% to 77.9\% while reducing latency from 1.90 to 0.87~s.
% In contrast, Moshi increased its TOR from 73.1\% to 79.8\% but showed a slightly longer latency in MT-QA (1.60 vs.\ 1.79~s), and the RL-based models also exhibited mixed changes across the two metrics.
% These differences suggest that turn-taking behavior depended on the dialogue settings.

We compared smooth turn-taking and user barge-in handling across Chat and MT-QA.
Most models achieved higher smooth turn-taking TORs in MT-QA.
For example, DuplexCascade increased from 69.6\% to 85.8\%, with mean latency decreasing from 4.92 to 2.99~s.
J-Moshi showed a larger TOR increase, from 31.5\% to 77.9\%, with latency decreasing from 1.90 to 0.87~s.
This pattern may reflect the clearer turn-taking structure of QA, where a question typically calls for an answer, whereas responses in Chat can be optional and turn transitions more ambiguous.
Models with already high Chat TORs generally showed smaller gains or slight decreases in MT-QA, consistent with less room for improvement.

Barge-in handling showed a different pattern.
PersonaPlex, PersonaPlex-RL, and both Japanese models had shorter mean stop latencies in MT-QA, whereas the remaining models had shorter latencies in Chat.
For example, PersonaPlex decreased from 2.06 to 1.28~s, while Moshi increased from 1.03 to 2.26~s.
For the former group, the onset of a new question may provide a clearer cue to yield the floor.
The opposite pattern in other models suggests that this cue is not used equally across systems.
These models may respond more readily to the prosodic and contextual cues in natural Chat interruptions than to those in synthetic MT-QA.

% The cross-lingual comparison revealed differences in both interaction timing and response content.
% In Chat, J-Moshi had low TORs for both smooth turn-taking and pause handling (31.5\% and 22.8\%, respectively).
% In contrast, LLM-jp-Moshi achieved a much higher smooth turn-taking TOR (84.0\%) but also a high pause-handling TOR (69.1\%).
Across languages, both Japanese models showed smaller separation between $\mathrm{SHIFT}$ and $\mathrm{PAUSE}$ TORs in Chat than the English Moshi-family models.
In Chat, LLM-jp-Moshi achieved a much better smooth turn-taking behavior~(higher smooth turn-taking TOR (84.0\%)) at the cost of poorer pause handling~(high pause-handling TOR (69.1\%)).
In contrast, J-Moshi had low TORs for both (31.5\% and 22.8\%).
Thus, the evaluated Japanese models showed less separation between responses to turn ends and user-held pauses.
They also tended to obtain lower MT-QA response quality scores than the English Moshi-based models.
For example, J-Moshi scored 49.7 in relevance, 63.9 in consistency, and 11.8 in QA score, compared with 77.7, 66.8, and 26.2 for PersonaPlex-RL.
\cref{fig:case_study_model} shows examples of English and Japanese model responses to the same question.

\begin{table}[!t]
    \begin{spacing}{0.75}
    \caption{Effect of dialogue context on MT-QA response quality. Gen.: \textit{Multi (gen)}; Ref.: \textit{Multi (ref)}. $+/-$ indicate improvement/degradation to Single.}
    \label{tab:context_response_quality}
    \end{spacing}
    \setlength{\tabcolsep}{1pt}
    \setlength{\aboverulesep}{1pt}
    \setlength{\belowrulesep}{1pt}
    \centering
    \small
    \renewcommand{\arraystretch}{0.95}
    \begin{tabular}{lccc@{}p{2pt}@{}ccc@{}p{2pt}@{}ccc}
    \toprule
    \multirow{2}{*}{\textbf{Model}} & \multicolumn{3}{c}{\textbf{Relevance}$\uparrow$} & & \multicolumn{3}{c}{\textbf{Consistency}$\uparrow$} & & \multicolumn{3}{c}{\textbf{QA Score}$\uparrow$} \\
    \cmidrule(l{\tabcolsep}r{\tabcolsep}){2-4} \cmidrule(l{\tabcolsep}r{\tabcolsep}){6-8} \cmidrule(l{\tabcolsep}r{\tabcolsep}){10-12}
    & Single & Gen. & Ref. & & Single & Gen. & Ref. & & Single & Gen. & Ref. \\
    \midrule
    Freeze-Omni & 62.6 & \degrade{6.8}{56.0} & -- && 61.4 & \degrade{8.1}{53.5} & -- && 16.7 & \improve{0.3}{17.0} & -- \\
    DuplexCascade & 63.0 & \degrade{1.8}{61.3} & -- && \textbf{70.2} & \degrade{10.3}{60.2} & -- && \textbf{19.8} & \improve{4.5}{24.2} & -- \\
    BayLing-Duplex & 71.3 & \degrade{9.7}{61.9} & -- && 62.4 & \degrade{11.3}{51.4} & -- && 19.6 & \degrade{2.2}{17.5} & -- \\
    Moshi & 60.6 & \improve{12.0}{72.3} & \improve{12.5}{72.8} && 54.1 & \improve{10.2}{64.0} & \improve{16.8}{70.5} && 11.1 & \improve{10.8}{21.6} & \improve{15.9}{26.6} \\
    Moshi-RL & 68.4 & \improve{9.8}{\textbf{77.9}} & \improve{3.6}{71.9} && 50.2 & \improve{15.3}{65.1} & \improve{20.0}{69.7} && 14.6 & \improve{9.7}{24.0} & \improve{12.5}{26.8} \\
    PersonaPlex & 76.9 & \degrade{0.4}{76.5} & \improve{5.3}{\textbf{82.0}} && 56.9 & \improve{11.0}{\textbf{67.6}} & \improve{17.5}{73.9} && 15.7 & \improve{10.1}{25.5} & \improve{17.8}{\textbf{33.0}} \\
    PersonaPlex-RL & \textbf{79.9} & \degrade{2.3}{77.7} & \improve{1.8}{81.6} && 59.5 & \improve{7.5}{66.8} & \improve{16.0}{\textbf{75.1}} && 17.9 & \improve{8.5}{\textbf{26.2}} & \improve{14.2}{31.7} \\
    \midrule\midrule
    J-Moshi & \textbf{58.0} & \degrade{8.5}{\textbf{49.7}} & \degrade{7.3}{50.9} && \textbf{63.0} & \improve{0.9}{\textbf{63.9}} & \degrade{2.3}{60.8} && \textbf{11.5} & \improve{0.3}{\textbf{11.8}} & \improve{3.2}{\textbf{14.6}} \\
    LLM-jp-Moshi & 52.2 & \degrade{7.0}{45.4} & \degrade{0.8}{51.4} && 61.0 & \degrade{3.6}{57.5} & \improve{0.3}{\textbf{61.3}} && 10.2 & \improve{0.3}{10.5} & \improve{4.2}{14.3} \\
    \bottomrule
    \end{tabular}
    \vspace{-2mm}
\end{table}

\begin{figure}[!t]
    \centering
    \includegraphics[width=1.0\linewidth]{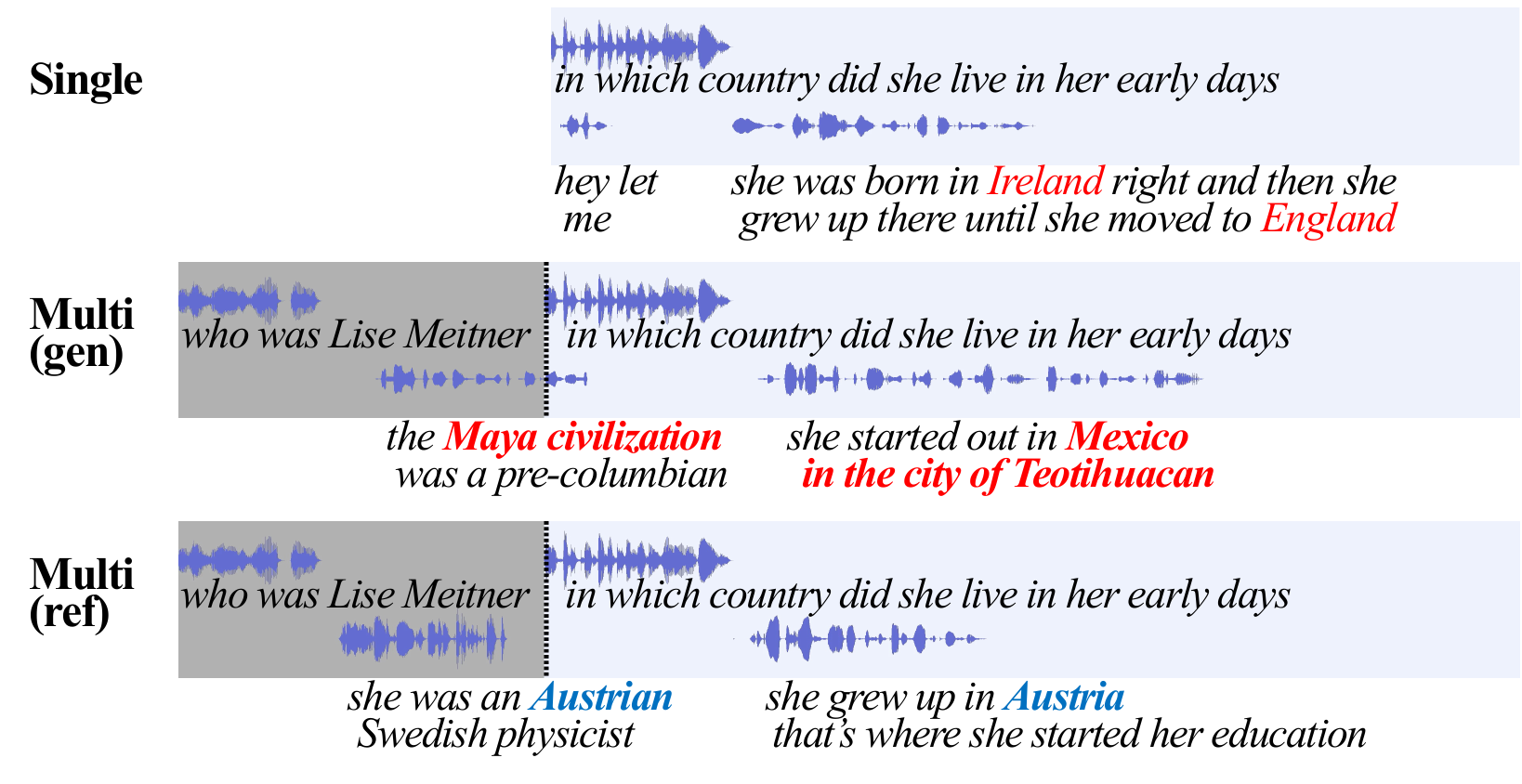}
    \vspace{-6mm}
    % \caption{Comparison of conditions in different contexts.}
    \caption{Case study showing the effect of dialogue context in MT-QA. Responses are obtained by PersonaPlex under the same context-dependent question using three context conditions. The gray region denotes the context region.}
    \label{fig:case_study_context}
    \vspace{-3mm}
\end{figure}

\subsection{ANALYSIS}
\subsubsection{Effect of Dialogue Context on Content}
We examined how dialogue context affects response content in MT-QA.
\cref{tab:context_response_quality} shows response relevance, context consistency, and QA score under the \textit{Single}, \textit{Multi (gen)}, and \textit{Multi (ref)} settings.
The effect of dialogue history varied across models.
For Freeze-Omni, DuplexCascade, and BayLing-Duplex, \textit{Multi (gen)} generally reduced relevance and consistency relative to \textit{Single}, while its effect on QA score was small or mixed.
In contrast, the English Moshi-based models generally benefited from multi-turn context, particularly in consistency and QA score.
For example, Moshi improved from 54.1 to 64.0 in consistency and from 11.1 to 21.6 in QA score from \textit{Single} to \textit{Multi (gen)}.
These results showed that access to dialogue history did not have a uniform effect across models.
Comparing \textit{Multi (gen)} with \textit{Multi (ref)} further revealed sensitivity to context mismatch.
For the English Moshi-based models, reference conditioning consistently increased context consistency and QA score.
For example, PersonaPlex improved from 67.6 to 73.9 in consistency and from 25.5 to 33.0 in QA score, while PersonaPlex-RL improved from 66.8 to 75.1 and from 26.2 to 31.7, respectively.
The Japanese models showed a different pattern.
For both J-Moshi and LLM-jp-Moshi, relevance decreased from \textit{Single} to \textit{Multi (gen)} and remained below the \textit{Single} score under \textit{Multi (ref)}.
Consistency changed only modestly, whereas QA score increased under \textit{Multi (ref)}, reaching 14.6 for J-Moshi and 14.3 for LLM-jp-Moshi.
\cref{fig:case_study_context} illustrates the effect of context mismatch qualitatively.
Under \textit{Single} without context, the model failed to answer the context-dependent question. Under \textit{Multi (gen)}, the generated history led to an inconsistent dialogue trajectory and an incorrect answer, whereas \textit{Multi (ref)} preserved the relevant preceding information and produced a contextually consistent answer.

\begin{figure}[!t]
    \centering
    \includegraphics[width=1.0\linewidth]{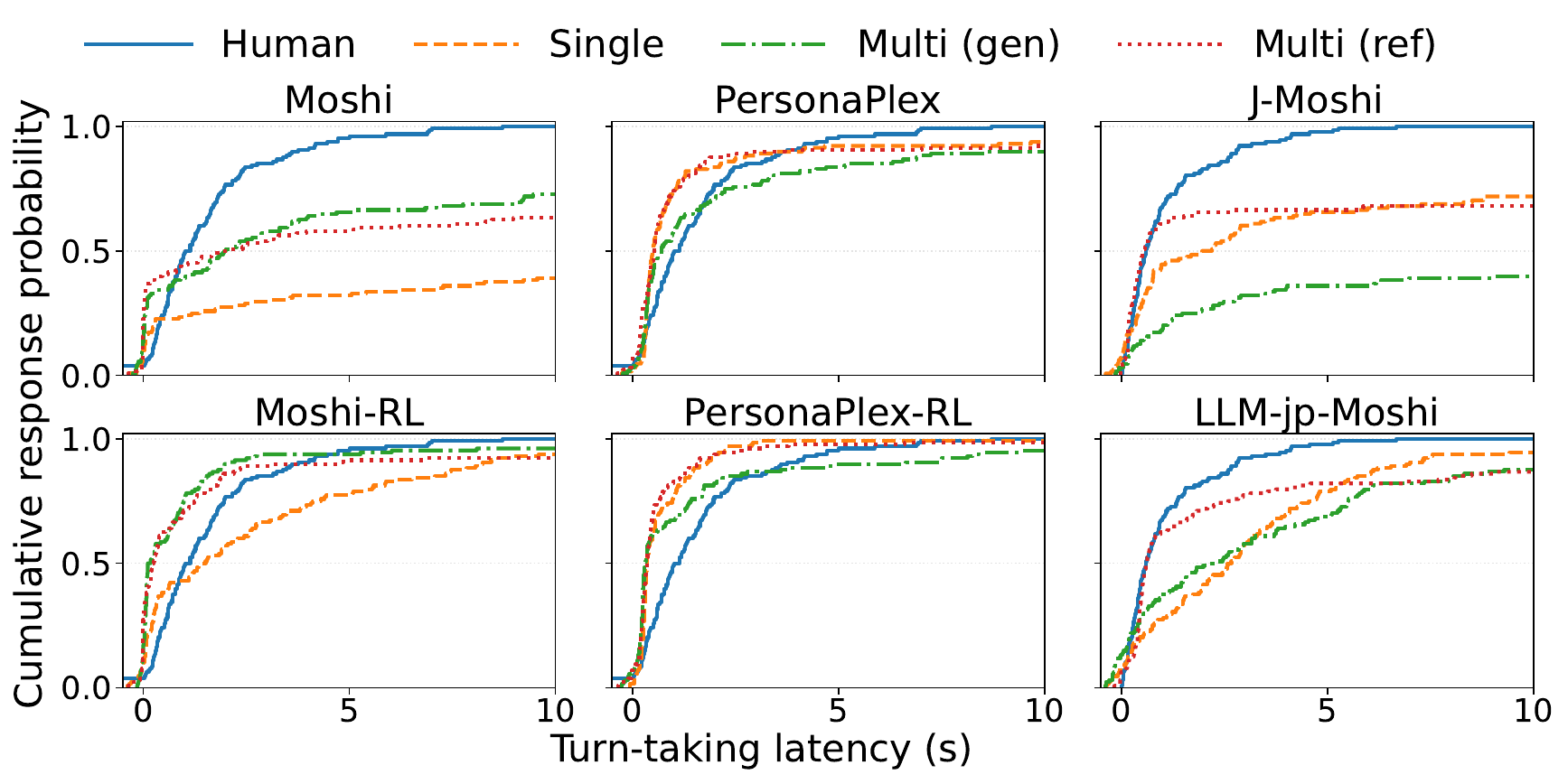}
    \vspace{-4mm}
    \caption{Cumulative probability distribution of response latency under different context settings.}
    \label{fig:context_effect_latency}
    \vspace{-3mm}
\end{figure}

\begin{figure}[!t]
    \centering
    \includegraphics[width=1.0\linewidth]{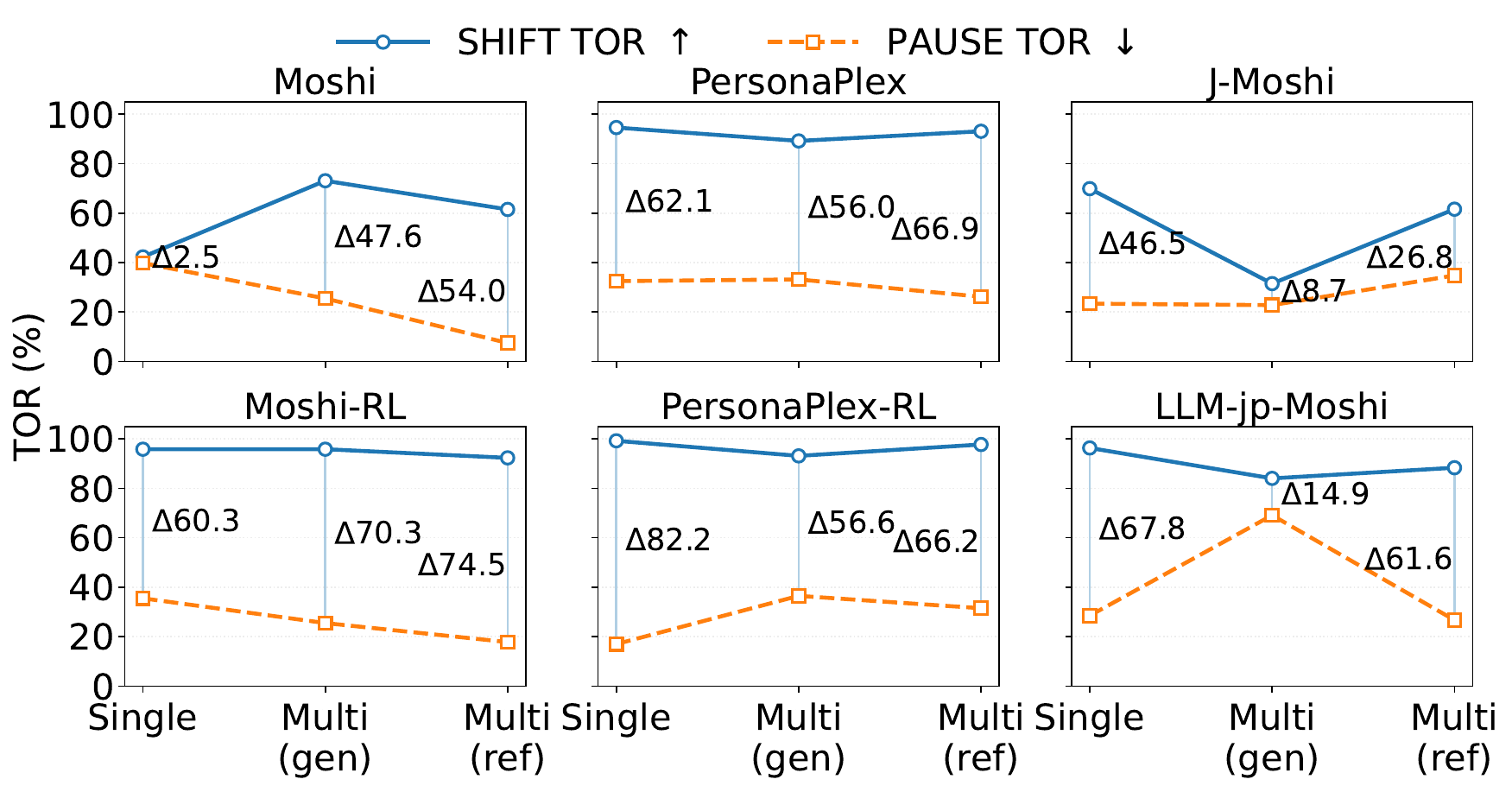}
    \vspace{-4mm}
    \caption{Effect of dialogue context on interaction timing. TORs for $\mathrm{SHIFT}$ and $\mathrm{PAUSE}$ events are shown under the \textit{Single}, \textit{Multi (gen)}, and \textit{Multi (ref)} settings. $\Delta$ indicates the TOR gap between $\mathrm{SHIFT}$ and $\mathrm{PAUSE}$.} %A larger $\Delta$ indicates a greater discrepancy between TORs of $\mathrm{SHIFT}$ and $\mathrm{PAUSE}$.}
    \label{fig:context_effect_timing}
    % \vspace{-3mm}
\end{figure}

\subsubsection{Effect of Dialogue Context on Timing}
We compared turn-taking behavior across \textit{Single}, \textit{Multi (gen)}, and \textit{Multi (ref)} on Chat for the six models supporting reference conditioning.
\cref{fig:context_effect_timing} shows the $\mathrm{SHIFT}$ and $\mathrm{PAUSE}$ TORs, with $\Delta$ denoting their difference.
A larger $\Delta$ indicates greater separation between responses to turn transitions and user-held pauses.
\cref{fig:context_effect_latency} shows the cumulative probability distribution of response latency.

We first compared \textit{Single} with \textit{Multi (ref)} to examine sensitivity to the presence of reference dialogue history.
In \cref{fig:context_effect_latency}, PersonaPlex and PersonaPlex-RL showed relatively similar response curves under these two conditions.
The other four models showed larger differences, with \textit{Multi (ref)} generally yielding higher early response probabilities than \textit{Single}.
Context effects on whether to take the turn also varied across models.
In \cref{fig:context_effect_timing}, $\Delta$ changed relatively little for PersonaPlex (62.1 to 66.9) and LLM-jp-Moshi (67.8 to 61.6), whereas larger changes were observed for Moshi (2.5 to 54.0) and J-Moshi (46.5 to 26.8).
These results indicate that dialogue context can affect both whether and how quickly models take the turn, with the effects varying across models.

We then examined the effect of the dialogue-history trajectory by comparing \textit{Multi (gen)} with \textit{Multi (ref)}.
In \cref{fig:context_effect_timing}, $\Delta$ was larger under \textit{Multi (ref)} for all models.
Differences were also observed in response latency.
For example, in \cref{fig:context_effect_latency}, PersonaPlex, PersonaPlex-RL, J-Moshi and LLM-jp-Moshi showed lower cumulative response probabilities under \textit{Multi (gen)} than under \textit{Multi (ref)}.
% These differences indicate sensitivity to the choice of system history and are consistent with a possible effect of context mismatch on timing.
These results were consistent with the possibility that context mismatch caused turn-taking latency to be overestimated in standard static multi-turn evaluation.

% Moshi's results also depended on initialization.
% Moshi tends to produce a greeting at the start of inference, which can overlap with the evaluated user utterance in \textit{Single}.
% To examine this behavior, we repeated the \textit{Single} evaluation with 5~s of silence prepended to the user-side input.
% Specifically, the model first processed 5~s of silent input while being allowed to generate speech, followed by the original user audio.
% This provided an opportunity for the initial greeting to occur before the user began speaking.
% In this diagnostic, which is not plotted, the $\mathrm{SHIFT}$ and $\mathrm{PAUSE}$ TORs were 66.5% and 20.0%, respectively, giving $\Delta=46.5$ percentage points.
% This was substantially larger than the original \textit{Single} value of 2.5 and close to the \textit{Multi (gen)} value of 47.6 in \cref{fig:context_effect_timing}.
% These results suggest that Moshi's differences between \textit{Single} and the multi-turn settings partly reflect initialization behavior.

These observations indicate that dialogue context can affect both response content and timing, with effects varying across models.
Evaluation under a single context condition may therefore provide an incomplete picture of model behavior.
\textit{Single} may underestimate the timing performance of models that benefit from preceding dialogue.
$\textit{Multi (gen)}$ may be affected by context mismatch, while \textit{Multi (ref)} uses teacher-forced history that may differ from the model's own generated trajectory.
These conditions provide complementary views of model behavior, and comparing them helps characterize both timing performance and sensitivity to dialogue history.

\section{LIMITATION}
\textit{Multi (ref)} uses teacher-forced reference histories that may differ from the model's own generated dialogue in content, timing, and speech characteristics.
Its results therefore characterize behavior under reference conditioning and may not fully reflect performance in free-running interaction.
Sensitivity to this conditioning may also vary across models.
The two dialogue settings also differ in data construction.
Chat is based on natural conversations, whereas MT-QA uses synthetically generated speech and timing, with translation additionally used for the Japanese MT-QA data.
The observed differences between Chat and MT-QA may therefore reflect both dialogue domain and data construction.
Finally, errors in automatic event detection, timing measurement, and model-based response-quality assessment may affect the reported results.

\section{CONCLUSION}
In this work, we presented M3-DuplexBench, a benchmark for evaluating FDSDSs across multiple turns, languages, and dialogue settings.
M3-DuplexBench jointly evaluates interaction timing and response content in English and Japanese, covering both real conversational speech and synthetic multi-turn question answering.
We further analyzed how dialogue history affected model behavior by comparing single-turn, free-running multi-turn, and reference-conditioned multi-turn settings.
Future work will extend M3-DuplexBench to broader languages, dialogue settings, and model architectures, and further investigate the relationship between dialogue history and interactive behavior.

% \section*{ACKNOWLEDGEMENT}
% This manuscript was edited and polished with the assistance of generative AI.
% All experimental design, implementation, and analysis were conducted by the authors who take full responsibility for the content.

% \section*{REFERENCES}
\bibliographystyle{IEEEtran}
\bibliography{custom}

\clearpage
\section*{APPENDIX}

\subsection{TIMING METRIC COMPUTATION}
\label{app:timing_metrics}

We compute timing metrics from timestamped segments on the model output channel.
For event $i$, let $s_i$ and $t_i$ denote the event onset and offset relative to the beginning of the evaluated input.
The generation window is $W_i=[s_i,s_i+\Delta)$, where $\Delta=10$~s.
Each model speech segment $k$ has an onset $a_{ik}$ and an offset $b_{ik}$.
We consider segments that overlap $W_i$, either partially or entirely, satisfying $b_{ik}>s_i$ and $a_{ik}<s_i+\Delta$.

\paragraph{Smooth turn-taking.}
For a SHIFT event, the response candidates are
\begin{equation}
    \mathcal{S}_i =
    \left\{
        k \mid a_{ik} \geq s_i-\epsilon
    \right\},
    \qquad \epsilon=0.4.
\end{equation}
Thus, a candidate must extend beyond the event onset and must not begin more than 0.4~s before it.
If multiple candidates exist, we select
\begin{equation}
    k_i^\star =
    \operatorname*{arg\,min}_{k\in\mathcal{S}_i}
    |a_{ik}-s_i|.
\end{equation}
The selected segment therefore has the onset closest to the event onset, which need not be the earliest candidate.
The event-level takeover indicator is $T_i^{\mathrm{SHIFT}}=1$ if $\mathcal{S}_i$ is nonempty and zero otherwise.
For a selected response, latency is $L_i^{\mathrm{SHIFT}}=a_{ik_i^\star}-s_i$.
Negative latency denotes an overlapping response.
If no candidate exists, the evaluator returns a zero takeover indicator and an undefined latency.
TOR is the percentage of evaluated samples with $T_i^{\mathrm{SHIFT}}=1$.
The reported latency is the mean of $L_i^{\mathrm{SHIFT}}$ over these samples.

\paragraph{Pause handling.}
For a PAUSE event, takeover is evaluated over the actual user pause $[s_i,t_i)$.
We exclude model output segments whose duration is less than 1.0~s and that contain fewer than four transcript chunks, treating them as backchannels.
Let $\mathcal{P}_i$ contain the remaining model segments that satisfy $a_{ik}<t_i$ and $b_{ik}>s_i$.
The takeover indicator is $T_i^{\mathrm{PAUSE}}=1$ if $\mathcal{P}_i$ is nonempty
and zero otherwise.

Latency is computed over
\begin{equation}
    \mathcal{F}_i =
    \mathcal{P}_i \cup \{k \mid a_{ik} \geq t_i\},
\end{equation}
which includes eligible segments overlapping the pause
and all segments starting at or after its offset.
The latency is
\begin{equation}
    L_i^{\mathrm{PAUSE}} =
    \begin{cases}
        \displaystyle \min_{k\in\mathcal{F}_i} a_{ik}-s_i,
        & \mathcal{F}_i\neq\emptyset,\\[4pt]
        \Delta,
        & \mathcal{F}_i=\emptyset.
    \end{cases}
\end{equation}

TOR is the percentage of evaluated samples with $T_i^{\mathrm{PAUSE}}=1$.
The reported latency is the mean of $L_i^{\mathrm{PAUSE}}$ over all evaluated samples.

\paragraph{User backchannels and barge-ins.}
BC and BARGE\_IN use the same segment-selection procedure
for stop latency.
With a short-pause tolerance of $\delta=0.5$~s, the
candidate set is
\begin{equation}
    \mathcal{E}_i =
    \left\{k\mid a_{ik}<s_i+\delta\right\}.
\end{equation}
If this set is nonempty, the segment with the earliest
onset is selected:
\begin{equation}
    j_i^\star =
    \operatorname*{arg\,min}_{k\in\mathcal{E}_i} a_{ik},
    \qquad
    L_i^{\mathrm{stop}}=b_{ij_i^\star}-s_i.
\end{equation}
Otherwise, stop latency is undefined.
For each event, the reported stop latency is the mean of $L_i^{\mathrm{stop}}$ over samples.

\subsection{LLM-BASED CONTENT EVALUATION}
\label{app:content_evaluation}
We use GPT-5 nano to evaluate the textual content of model responses to SHIFT events.
Response relevance and context consistency are evaluated in both Chat and MT-QA, whereas QA score is evaluated only in MT-QA.
Each metric uses a separate system prompt defining the evaluation criteria and a user message containing the text to be evaluated.
The judge is instructed to return only an integer score $r_i\in\{0,1,2\}$.
For each metric, we report the mean score normalized to a 0--100 scale.
The system prompts and user-message templates are provided below.
Braced expressions denote fields replaced with the corresponding sample text.

\paragraph{Response relevance.}
This metric evaluates whether the response addresses
the current user utterance, independently of factual
correctness.

\noindent\textit{System prompt:}
\begin{quote}
\footnotesize
\raggedright
Judge whether the model response is relevant to the
current user utterance.
Output only one integer: 0, 1, or 2.\par
2: relevant and responsive.\par
1: partially relevant, vague, generic, or incomplete.\par
0: irrelevant, unrelated, or not a meaningful response.\par
Judge only relevance to the current user utterance,
not factual correctness.
\end{quote}

\noindent\textit{User-message template:}
\begin{quote}
\footnotesize
\raggedright
User Utterance: \texttt{\{user\_utterance\}}\par
Model Response: \texttt{\{model\_answer\}}\par
Your output (0, 1, or 2):
\end{quote}

\paragraph{Context consistency.}
This metric evaluates agreement with previous statements and established information in the supplied dialogue context, independently of factual accuracy and relevance to the current user utterance.
The judge receives the reference dialogue history as \texttt{dialogue\_context}.

\noindent\textit{System prompt:}
\begin{quote}
\footnotesize
\raggedright
Judge whether the model response is consistent with its
own previous statements in the dialogue context.
Output only one integer: 0, 1, or 2.\par
2: consistent and grounded in the previous dialogue
context.\par
1: not clearly grounded in the context, but does not
contradict it.\par
0: contradicts the model's previous statements or
established context.\par
Judge only consistency with the dialogue context,
not factual accuracy or response relevance to the
current user utterance.
\end{quote}

\noindent\textit{User-message template:}
\begin{quote}
\footnotesize
\raggedright
Dialogue Context:\par
\texttt{\{dialogue\_context\}}\par
Model Response: \texttt{\{model\_response\}}\par
Your output (0, 1, or 2):
\end{quote}

\paragraph{QA score.}
This metric evaluates the answer against a reference
answer, allowing partial credit for incomplete or
partially correct responses.

\noindent\textit{System prompt:}
\begin{quote}
\footnotesize
\raggedright
Judge whether the model response is correct with respect
to the reference answer.
Output only one integer: 0, 1, or 2.\par
2: correct or semantically equivalent to the reference
answer.\par
1: partially correct, incomplete, vague, or contains
some correct information.\par
0: incorrect, unsupported by the reference answer,
or not an answer to the user utterance.\par
Judge factual accuracy against the reference answer,
not style or fluency.
\end{quote}

\noindent\textit{User-message template:}
\begin{quote}
\footnotesize
\raggedright
User Question: \texttt{\{user\_question\}}\par
Model Answer: \texttt{\{model\_answer\}}\par
Reference Answer: \texttt{\{reference\_answer\}}\par
Your output (0, 1, or 2):
\end{quote}

\subsection{INFERENCE SETUP}
Inference was performed on NVIDIA RTX A6000 GPUs.
Freeze-Omni and DuplexCascade used a client--server architecture, with the client and server running on the same machine.
The Freeze-Omni server used a single GPU, whereas DuplexCascade used two GPUs: one for ASR and TTS, and the other for LLM inference.
All remaining models were evaluated using offline inference.
We used the default inference parameters for each model.

\begin{table}[t]
    \centering
    \begin{spacing}{0.75}
    \caption{Moshi results under the Single condition in English Chat, with and without 5~s of initial silence.}
    \label{tab:moshi_initial_silence}
    \end{spacing}
    \footnotesize
    \setlength{\aboverulesep}{1pt}
    \setlength{\belowrulesep}{1pt}
    \centering
    \footnotesize
    \renewcommand{\arraystretch}{0.9}
    \begin{tabular}{lll}
        \toprule
        \textbf{Metric} & \textbf{No prepend} & \textbf{+5~s silence} \\
        \midrule
        SHIFT TOR (\%) $\uparrow$ & 42.3{\tiny$\pm$6.0} & 88.8{\tiny$\pm$3.9} \\
        SHIFT latency (s) $\downarrow$ & 2.14{\tiny$\pm$0.56} & 2.94{\tiny$\pm$0.42} \\
        PAUSE TOR (\%) $\downarrow$ & 39.8{\tiny$\pm$4.8} & 18.8{\tiny$\pm$3.8} \\
        PAUSE latency (s) $\uparrow$ & 4.57{\tiny$\pm$0.48} & 5.12{\tiny$\pm$0.38} \\
        Relevance $\uparrow$ & 62.3{\tiny$\pm$2.9} & 77.1{\tiny$\pm$3.2} \\
        Consistency $\uparrow$ & 63.9{\tiny$\pm$3.2} & 83.5{\tiny$\pm$3.0} \\
        \bottomrule
    \end{tabular}
    \vspace{-2mm}
\end{table}

\subsection{EFFECT OF INITIAL SILENCE ON MOSHI}
Moshi showed low smooth turn-taking TOR and response-quality
scores under the Single condition.
One possible explanation is its tendency to initiate a conversation with a greeting rather than wait for the user.
At a cold start, this greeting may overlap with the user utterance and affect the subsequent response.
To examine sensitivity to dialogue initialization, we prepended 5~s of silence to the user input and evaluated Moshi under the Single condition in English Chat.
As shown in \cref{tab:moshi_initial_silence}, initial silence increased SHIFT TOR from 42.3\% to 88.8\% and reduced PAUSE TOR from 39.8\% to 18.8\%.
Relevance and consistency also increased.
The results suggest that a model's conversation-initiation strategy can affect both timing and content scores when user inputs follow a fixed schedule.

\begin{table}[t]
    \centering
    \begin{spacing}{0.75}
    \caption{Comparison of timing evaluation algorithms on the same English Chat outputs under the Single condition. M3 denotes M3-DuplexBench, and FDB denotes our implementation of FDB-style scoring. TOR is reported in percent and latency in seconds.}
    \label{tab:metric_comparison}
    \end{spacing}
    \small
    \setlength{\tabcolsep}{2.2pt}
    \setlength{\aboverulesep}{1pt}
    \setlength{\belowrulesep}{1pt}
    \centering
    \footnotesize
    \renewcommand{\arraystretch}{0.9}    \begin{tabular}{llccc}
        \toprule
        \multirow{2}{*}{\textbf{Model}} & \multirow{2}{*}{\textbf{Scoring}} & \multicolumn{2}{c}{\textit{Smooth turn-taking}}
        & \textit{Pause Handling} \\
        \cmidrule(lr){3-4} \cmidrule(lr){5-5}
        &  & TOR$\uparrow$
        & Latency (s)$\downarrow$ & TOR$\downarrow$ \\        \midrule
        Moshi & M3 & 42.3{\tiny$\pm$6.0} & 2.14{\tiny$\pm$0.56} & 39.8{\tiny$\pm$4.8} \\
        & FDB & 95.0{\tiny$\pm$2.7} & 0.00{\tiny$\pm$nan} & 93.0{\tiny$\pm$2.5} \\ \midrule
        PersonaPlex & M3 & 94.6{\tiny$\pm$2.8} & 0.78{\tiny$\pm$0.14} & 32.5{\tiny$\pm$4.6} \\
        & FDB & 94.2{\tiny$\pm$2.9} & 0.50{\tiny$\pm$0.08} & 79.7{\tiny$\pm$4.0} \\ \bottomrule
    \end{tabular}
    \vspace{-4mm}
\end{table}

\subsection{SENSITIVITY TO TIMING METRIC DEFINITIONS}
Although our timing metrics build on Full-Duplex-Bench (FDB), the observed scores differ considerably from previously reported results.
For example, Roy et al.~\cite{Roy2026-db} report smooth turn-taking and pause-handling TORs of 94.1\% and 98.0\%, respectively, for Moshi on Candor, whereas our corresponding single-turn Chat scores are 42.3\% and 39.8\%.
Differences in data selection and event definitions may contribute to this discrepancy, but the metric definitions themselves can also affect the results.
For smooth turn-taking, FDB evaluates the output as a whole, excludes short outputs, and measures latency
from the first transcript chunk, clipping negative values to zero.
Our evaluator instead selects the eligible segment whose onset is closest to the turn end, permits at most 0.4~s of anticipatory overlap, retains negative latencies, and accepts short responses.

For pause handling, M3-DuplexBench counts a takeover when non-backchannel speech overlaps the actual user pause.
FDB instead counts a takeover when speech occurs anywhere in the supplied output, unless the output as a whole is classified as a short utterance.

To examine this sensitivity, we scored the same single-turn outputs from Moshi and PersonaPlex in
English Chat using both algorithms (\cref{tab:metric_comparison}).
With FDB-style scoring, Moshi's smooth turn-taking TOR increased from 42.3\% to 95.0\%, while its mean latency decreased from 2.135 to 0.000~s.
This pattern is consistent with counting early speech as takeover and assigning it zero latency.
PersonaPlex's turn-taking TOR changed little, but its mean latency also decreased.
Pause-handling TOR increased substantially for both models.
These results demonstrate that metric definitions can substantially change both absolute scores and model
comparisons.

\begin{table*}[t]
    \caption{Detailed results on the chat domain. Values are reported as mean$\pm$95\% confidence interval.~Single = no dialogue context;  Multi (gen) = user-side context only; Multi (ref) = user- and system-side context.}
    % \vspace{-2mm}
    \setlength{\tabcolsep}{2.2pt}
    \setlength{\aboverulesep}{1pt}
    \setlength{\belowrulesep}{1pt}
    \centering
    \renewcommand{\arraystretch}{0.9}
    \footnotesize
    \begin{tabular}{lllcccccccc} \toprule
        & \multirow{3}{*}{\textbf{Model}} & \multirow{3}{*}{\textbf{Context}} & \multicolumn{2}{c}{\textit{Smooth}} & \multicolumn{2}{c}{\textit{Pause}} & \textit{User Back-} & \textit{User} & \multicolumn{2}{c}{\multirow{2}{*}{\textit{Response Quality}}}  \\
        & & & \multicolumn{2}{c}{\textit{Turn Taking}} & \multicolumn{2}{c}{\textit{Handling}} & \textit{channeling} & \textit{Barge-in} &  \\
        \cmidrule(l{\tabcolsep}r{\tabcolsep}){4-5} \cmidrule(l{\tabcolsep}r{\tabcolsep}){6-7} \cmidrule(l{\tabcolsep}r{\tabcolsep}){8-8} \cmidrule(l{\tabcolsep}r{\tabcolsep}){9-9} \cmidrule(l{\tabcolsep}r{\tabcolsep}){10-11}
        & & & TOR$\uparrow$ & Latency~(s)$\downarrow$ & TOR$\downarrow$ & Latency~(s)$\uparrow$ & Stop Latency~(s)$\uparrow$ & Stop Latency~(s)$\downarrow$ & Relevance$\uparrow$ & Consistency$\uparrow$ \\ \midrule
        \multirow{20}{*}{\rotatebox{90}{\textbf{English Chat}}}
        & DuplexCascade & Single & 72.7{\tiny$\pm$5.5} & 4.70{\tiny$\pm$0.42} & 0.3{\tiny$\pm$0.5} & 6.91{\tiny$\pm$0.30} & -- & -- & 61.3{\tiny$\pm$2.6} & 77.7{\tiny$\pm$3.2} \\
        &               & Multi (gen) & 69.6{\tiny$\pm$5.6} & 4.92{\tiny$\pm$0.43} & 0.3{\tiny$\pm$0.5} & 7.58{\tiny$\pm$0.28} & 0.41{\tiny$\pm$0.25} & 0.35{\tiny$\pm$0.21} & 64.8{\tiny$\pm$2.9} & 77.7{\tiny$\pm$3.2} \\ \cmidrule(l{\tabcolsep}r{\tabcolsep}){3-11}
        & Freeze-Omni & Single & 82.7{\tiny$\pm$4.6} & 3.06{\tiny$\pm$0.35} & 12.5{\tiny$\pm$3.3} & 5.55{\tiny$\pm$0.38} & -- & -- & 70.4{\tiny$\pm$3.2} & 80.2{\tiny$\pm$3.2} \\
        &             & Multi (gen) & 80.0{\tiny$\pm$4.9} & 2.66{\tiny$\pm$0.30} & 61.8{\tiny$\pm$4.8} & 1.90{\tiny$\pm$0.48} & 2.21{\tiny$\pm$0.45} & 1.71{\tiny$\pm$0.39} & 59.4{\tiny$\pm$2.9} & 68.7{\tiny$\pm$3.3} \\ \cmidrule(l{\tabcolsep}r{\tabcolsep}){3-11}
        & BayLing-Duplex & Single & 85.4{\tiny$\pm$4.3} & 2.49{\tiny$\pm$0.17} & 14.8{\tiny$\pm$3.5} & 3.66{\tiny$\pm$0.29} & -- & -- & 72.5{\tiny$\pm$3.6} & 59.4{\tiny$\pm$4.0} \\
        &                & Multi (gen) & 88.8{\tiny$\pm$3.9} & 1.67{\tiny$\pm$0.14} & 33.3{\tiny$\pm$4.6} & 2.05{\tiny$\pm$0.27} & 1.68{\tiny$\pm$0.59} & 1.69{\tiny$\pm$0.58} & 70.0{\tiny$\pm$3.4} & 60.4{\tiny$\pm$3.7}\\ \cmidrule(l{\tabcolsep}r{\tabcolsep}){3-11}
        % & Moshi\dag & Single & 88.8{\tiny$\pm$3.9} & 2.94{\tiny$\pm$0.42} & 18.8{\tiny$\pm$3.8} & 5.12{\tiny$\pm$0.38} & -- & -- & 77.1{\tiny$\pm$3.2} & 83.5{\tiny$\pm$3.0}\\
        & Moshi & Single & 42.3{\tiny$\pm$6.0} & 2.14{\tiny$\pm$0.56} & 39.8{\tiny$\pm$4.8} & 4.57{\tiny$\pm$0.48} & -- & -- & 62.3{\tiny$\pm$2.9} & 63.9{\tiny$\pm$3.2}\\
        & & Multi (gen) & 73.1{\tiny$\pm$5.4} & 1.60{\tiny$\pm$0.35} & 25.5{\tiny$\pm$4.3} & 4.82{\tiny$\pm$0.42} & 1.84{\tiny$\pm$0.63} & 1.03{\tiny$\pm$0.22} & 77.7{\tiny$\pm$3.3} & 88.3{\tiny$\pm$2.9} \\
        & & Multi (ref) & 61.5{\tiny$\pm$6.0} & 1.11{\tiny$\pm$0.31} & 7.5{\tiny$\pm$2.6} & 7.34{\tiny$\pm$0.36} & 3.04{\tiny$\pm$0.57} & 1.75{\tiny$\pm$0.38} & 66.9{\tiny$\pm$3.2} & 90.6{\tiny$\pm$2.4} \\ \cmidrule(l{\tabcolsep}r{\tabcolsep}){3-11} 
        & Moshi-RL & Single & 95.8{\tiny$\pm$2.5} & 2.15{\tiny$\pm$0.32} & 35.5{\tiny$\pm$4.7} & 1.73{\tiny$\pm$0.23} & -- & -- & 78.3{\tiny$\pm$3.1} & 68.3{\tiny$\pm$3.3} \\
        &  & Multi (gen) & 95.8{\tiny$\pm$2.5}  & 0.58{\tiny$\pm$0.15} & 25.5{\tiny$\pm$4.3} & 2.22{\tiny$\pm$0.24} & 1.25{\tiny$\pm$0.35} & 1.12{\tiny$\pm$0.22} & 82.7{\tiny$\pm$3.1} & 93.7{\tiny$\pm$2.1} \\
        &  & Multi (ref) & 92.3{\tiny$\pm$3.3} & 0.72{\tiny$\pm$0.17} & 17.8{\tiny$\pm$3.8} & 2.76{\tiny$\pm$0.26} & 2.41{\tiny$\pm$0.50} & 1.07{\tiny$\pm$0.19} & 72.3{\tiny$\pm$3.3} & 94.8{\tiny$\pm$2.0} \\ \cmidrule(l{\tabcolsep}r{\tabcolsep}){3-11}
        & PersonaPlex & Single & 94.6{\tiny$\pm$2.8} & 0.78{\tiny$\pm$0.14} & 32.5{\tiny$\pm$4.6} & 3.29{\tiny$\pm$0.35} & -- & -- & 70.8{\tiny$\pm$3.4}  & 72.3{\tiny$\pm$3.3} \\
        & & Multi (gen) & 89.2{\tiny$\pm$3.8} & 1.41{\tiny$\pm$0.25} & 33.3{\tiny$\pm$4.6} & 3.00{\tiny$\pm$0.38} & 4.07{\tiny$\pm$0.85} & 2.06{\tiny$\pm$0.63} & 66.7{\tiny$\pm$3.2} & 79.8{\tiny$\pm$3.2} \\
        & & Multi (ref) & 93.1{\tiny$\pm$3.1} & 0.65{\tiny$\pm$0.12} & 26.3{\tiny$\pm$4.3} & 3.76{\tiny$\pm$0.36} & 5.31{\tiny$\pm$0.71} & 2.30{\tiny$\pm$0.52} & 75.8{\tiny$\pm$3.1} & 89.8{\tiny$\pm$2.5} \\ \cmidrule(l{\tabcolsep}r{\tabcolsep}){3-11}
        & PersonaPlex-RL & Single & 99.2{\tiny$\pm$1.1} & 0.59{\tiny$\pm$0.07} & 17.0{\tiny$\pm$3.7} & 2.40{\tiny$\pm$0.17} & -- & -- & 77.1{\tiny$\pm$3.2} & 78.7{\tiny$\pm$3.2}  \\
        &                & Multi (gen) & 93.1{\tiny$\pm$3.1} & 0.97{\tiny$\pm$0.24} & 36.5{\tiny$\pm$4.7} & 1.61{\tiny$\pm$0.20} & 4.19{\tiny$\pm$0.89} & 1.86{\tiny$\pm$0.54} & 72.9{\tiny$\pm$3.2} & 86.9{\tiny$\pm$2.7} \\ 
        &                & Multi (ref) & 97.7{\tiny$\pm$1.8} & 0.62{\tiny$\pm$0.14} & 31.5{\tiny$\pm$4.6} & 2.27{\tiny$\pm$0.24} & 4.67{\tiny$\pm$0.63} & 2.00{\tiny$\pm$0.46} & 77.5{\tiny$\pm$3.2} & 89.2{\tiny$\pm$2.5} \\ \midrule \midrule
        \multirow{7}{*}{\rotatebox{90}{\textbf{Japanese Chat}}}
        & J-Moshi & Single & 69.9{\tiny$\pm$4.8} & 1.65{\tiny$\pm$0.25} & 23.4{\tiny$\pm$4.6} & 5.82{\tiny$\pm$0.46} & -- & -- & 71.2{\tiny$\pm$3.2} & 88.4{\tiny$\pm$2.4} \\
        &         & Multi (gen) & 31.5{\tiny$\pm$4.9} & 1.90{\tiny$\pm$0.47} & 22.8{\tiny$\pm$4.5} & 7.02{\tiny$\pm$0.50} & 2.51{\tiny$\pm$1.23} & 2.35{\tiny$\pm$0.52} & 61.2{\tiny$\pm$3.7} & 91.7{\tiny$\pm$2.0} \\
        &         & Multi (ref) & 61.6{\tiny$\pm$5.1} & 0.88{\tiny$\pm$0.22} & 34.8{\tiny$\pm$5.1} & 5.06{\tiny$\pm$0.50} & 1.98{\tiny$\pm$0.63} & 1.25{\tiny$\pm$0.28} & 63.0{\tiny$\pm$2.8} & 87.1{\tiny$\pm$2.5} \\ \cmidrule(l{\tabcolsep}r{\tabcolsep}){3-11}
        & LLM-jp-Moshi & Single & 96.3{\tiny$\pm$2.0} & 2.81{\tiny$\pm$0.25} & 28.5{\tiny$\pm$4.9} & 3.36{\tiny$\pm$0.31} & -- & -- & 59.9{\tiny$\pm$2.3} & 79.4{\tiny$\pm$2.7} \\
        &              & Multi (gen) & 84.0{\tiny$\pm$3.9} & 2.26{\tiny$\pm$0.29} & 69.1{\tiny$\pm$5.0} & 0.87{\tiny$\pm$0.36} & 3.07{\tiny$\pm$0.64} & 2.73{\tiny$\pm$0.37} & 59.7{\tiny$\pm$2.5} & 81.7{\tiny$\pm$2.6} \\
        &              & Multi (ref) & 88.3{\tiny$\pm$3.4} & 1.34{\tiny$\pm$0.21} & 26.7{\tiny$\pm$4.8} & 4.50{\tiny$\pm$0.39} & 2.25{\tiny$\pm$0.53} & 1.86{\tiny$\pm$0.32} & 62.2{\tiny$\pm$2.7} & 90.0{\tiny$\pm$2.2} \\
        \bottomrule
    \end{tabular}
    \label{tab:all_chat_results}
    % \vspace{-4mm}
\end{table*}

\begin{table*}[t]
    \caption{Detailed results on MT-QA domain. Values are reported as mean$\pm$95\% confidence interval.~Single = no dialogue context;  Multi (gen) = user-side context only; Multi (ref) = user- and system-side context.}
    % \vspace{-2mm}
    \setlength{\tabcolsep}{2.2pt}
    \setlength{\aboverulesep}{1pt}
    \setlength{\belowrulesep}{1pt}
    \centering
    \footnotesize
    \renewcommand{\arraystretch}{0.9}
    \begin{tabular}{lllccccccc} \toprule
        & \multirow{3}{*}{\textbf{Model}} & \multirow{3}{*}{\textbf{Context}} & \multicolumn{2}{c}{\textit{Smooth}} &  \textit{User} & \multicolumn{3}{c}{\multirow{2}{*}{\textit{Response Quality}}}  \\
        & & & \multicolumn{2}{c}{\textit{Turn Taking}} & \textit{Barge-in} &  \\
        \cmidrule(l{\tabcolsep}r{\tabcolsep}){4-5} \cmidrule(l{\tabcolsep}r{\tabcolsep}){6-6} \cmidrule(l{\tabcolsep}r{\tabcolsep}){7-9}
        & & & TOR$\uparrow$ & Latency~(s)$\downarrow$ & Stop Latency~(s)$\downarrow$ & Relevance$\uparrow$ & Consistency$\uparrow$ & QA Score$\uparrow$ \\ \midrule
        \multirow{20}{*}{\rotatebox{90}{\textbf{English MT-QA}}}
        & DuplexCascade & Single & 92.4{\tiny$\pm$2.3} & 2.78{\tiny$\pm$0.10} & -- & 63.0{\tiny$\pm$2.2} & 70.2{\tiny$\pm$2.4} & 19.8{\tiny$\pm$2.6} \\
        &               & Multi (gen) & 85.8{\tiny$\pm$3.0} & 2.99{\tiny$\pm$0.17} & 2.91{\tiny$\pm$0.41} & 61.3{\tiny$\pm$2.2} & 60.2{\tiny$\pm$2.9} & 24.2{\tiny$\pm$2.5} \\ \cmidrule(l{\tabcolsep}r{\tabcolsep}){3-9}
        & Freeze-Omni & Single & 90.5{\tiny$\pm$2.5} & 1.71{\tiny$\pm$0.13} & -- & 62.6{\tiny$\pm$2.4} & 61.4{\tiny$\pm$2.9} & 16.7{\tiny$\pm$2.5} \\
        &             & Multi (gen) & 85.4{\tiny$\pm$3.0} & 2.24{\tiny$\pm$0.19} & 2.43{\tiny$\pm$0.26} & 56.0{\tiny$\pm$2.3} & 53.5{\tiny$\pm$2.6} & 17.0{\tiny$\pm$2.7} \\ \cmidrule(l{\tabcolsep}r{\tabcolsep}){3-9}
        & BayLing-Duplex & Single & 87.5{\tiny$\pm$2.8} & 2.91{\tiny$\pm$0.14} & -- & 71.3{\tiny$\pm$2.4} & 62.4{\tiny$\pm$2.8} & 19.6{\tiny$\pm$2.6} \\
        &                & Multi (gen) & 91.7{\tiny$\pm$2.4} & 1.89{\tiny$\pm$0.12} & 1.76{\tiny$\pm$0.28}  & 61.9{\tiny$\pm$2.2} & 51.4{\tiny$\pm$2.9} & 17.5{\tiny$\pm$2.6} \\ \cmidrule(l{\tabcolsep}r{\tabcolsep}){3-9}
        & Moshi & Single & 41.8{\tiny$\pm$4.2} & 2.46{\tiny$\pm$0.39} & -- & 60.6{\tiny$\pm$2.1} & 54.1{\tiny$\pm$2.2} & 11.1{\tiny$\pm$2.2} \\
        &       & Multi (gen) & 79.8{\tiny$\pm$3.4} & 1.79{\tiny$\pm$0.25} & 2.26{\tiny$\pm$0.27}  & 72.3{\tiny$\pm$2.3} & 64.0{\tiny$\pm$2.8} & 21.6{\tiny$\pm$2.8} \\
        &       & Multi (ref) & 91.9{\tiny$\pm$2.3} & 0.41{\tiny$\pm$0.09} & 2.23{\tiny$\pm$0.26} & 72.8{\tiny$\pm$2.4} & 70.5{\tiny$\pm$2.6} & 26.6{\tiny$\pm$3.0} \\ \cmidrule(l{\tabcolsep}r{\tabcolsep}){3-9} 
        & Moshi-RL & Single & 97.0{\tiny$\pm$1.5} & 2.44{\tiny$\pm$0.22} & -- & 68.4{\tiny$\pm$2.3} & 50.2{\tiny$\pm$2.1} & 14.6{\tiny$\pm$2.4} \\
        &          & Multi (gen) & 94.3{\tiny$\pm$2.0} & 1.02{\tiny$\pm$0.18} & 1.95{\tiny$\pm$0.24} & 77.9{\tiny$\pm$2.3} & 65.1{\tiny$\pm$2.7} & 24.0{\tiny$\pm$2.8} \\ 
        &          & Multi (ref) & 94.1{\tiny$\pm$2.0} & 0.34{\tiny$\pm$0.05} & 1.82{\tiny$\pm$0.24} & 71.9{\tiny$\pm$2.5} & 69.7{\tiny$\pm$2.5} & 26.8{\tiny$\pm$2.9} \\ \cmidrule(l{\tabcolsep}r{\tabcolsep}){3-9}
        & PersonaPlex & Single & 97.2{\tiny$\pm$1.4} & 0.38{\tiny$\pm$0.03} & -- & 76.9{\tiny$\pm$2.4} & 56.9{\tiny$\pm$2.9} & 15.7{\tiny$\pm$2.3} \\
        &             & Multi (gen) & 90.2{\tiny$\pm$2.5} & 0.35{\tiny$\pm$0.02} & 1.28{\tiny$\pm$0.16}  & 76.5{\tiny$\pm$2.3} & 67.6{\tiny$\pm$2.7} & 25.5{\tiny$\pm$2.7} \\
        &             & Multi (ref) & 96.4{\tiny$\pm$1.6} & 0.28{\tiny$\pm$0.01} & 1.05{\tiny$\pm$0.13}  & 82.0{\tiny$\pm$2.2} & 73.9{\tiny$\pm$2.6} & 33.0{\tiny$\pm$3.1} \\ \cmidrule(l{\tabcolsep}r{\tabcolsep}){3-9}
        & PersonaPlex-RL & Single & 97.9{\tiny$\pm$1.2} & 0.31{\tiny$\pm$0.01} & -- & 79.9{\tiny$\pm$2.3} & 59.5{\tiny$\pm$2.9} & 17.9{\tiny$\pm$2.5} \\
        &                & Multi (gen) & 90.7{\tiny$\pm$2.5} & 0.32{\tiny$\pm$0.05} & 1.50{\tiny$\pm$0.20} & 77.7{\tiny$\pm$2.3} & 66.8{\tiny$\pm$2.7} & 26.2{\tiny$\pm$2.8} \\ 
        &                & Multi (ref) & 97.9{\tiny$\pm$1.2} & 0.27{\tiny$\pm$0.03} & 1.19{\tiny$\pm$0.16} & 81.6{\tiny$\pm$2.2} & 75.1{\tiny$\pm$2.5} & 31.7{\tiny$\pm$3.1} \\ \midrule \midrule
        \multirow{6}{*}{\rotatebox{90}{\textbf{Japanese MT-QA~}}}
        & J-Moshi & Single & 90.8{\tiny$\pm$2.5} & 1.08{\tiny$\pm$0.14} & -- & 58.0{\tiny$\pm$2.2} & 63.0{\tiny$\pm$2.5} & 11.5{\tiny$\pm$2.2} \\
        &         & Multi (gen) & 77.9{\tiny$\pm$3.6} & 0.87{\tiny$\pm$0.15} & 2.04{\tiny$\pm$0.27} & 49.7{\tiny$\pm$2.0} & 63.9{\tiny$\pm$2.5} & 11.8{\tiny$\pm$2.2} \\
        &         & Multi (ref) & 90.6{\tiny$\pm$2.6} & 0.31{\tiny$\pm$0.04} & 1.42{\tiny$\pm$0.19} & 50.9{\tiny$\pm$1.8} & 60.8{\tiny$\pm$2.3} & 14.6{\tiny$\pm$2.3} \\ \cmidrule(l{\tabcolsep}r{\tabcolsep}){3-9}
        & LLM-jp-Moshi & Single & 92.4{\tiny$\pm$2.3} & 2.76{\tiny$\pm$0.20} & -- & 52.2{\tiny$\pm$1.8} & 61.0{\tiny$\pm$2.4} & 10.2{\tiny$\pm$1.9} \\
        &              & Multi (gen) & 84.1{\tiny$\pm$3.2} & 1.87{\tiny$\pm$0.22} & 2.33{\tiny$\pm$0.28} & 45.4{\tiny$\pm$2.0} & 57.5{\tiny$\pm$2.3} & 10.5{\tiny$\pm$2.0} \\
        &              & Multi (ref) & 90.4{\tiny$\pm$2.6} & 0.69{\tiny$\pm$0.12} & 1.59{\tiny$\pm$0.21} & 51.4{\tiny$\pm$1.7} & 61.3{\tiny$\pm$2.4} & 14.3{\tiny$\pm$2.3} \\
        \bottomrule
    \end{tabular}
    \label{tab:all_qa_results}
    % \vspace{-4mm}
\end{table*}

\subsection{DETAILED BENCHMARK RESULTS}
\cref{tab:all_chat_results,tab:all_qa_results} present detailed results for Chat and MT-QA, respectively.

\end{document}